\documentclass{article} % For LaTeX2e
\usepackage[utf8]{inputenc} % allow utf-8 input
\usepackage[T1]{fontenc}
\usepackage{hyperref}  
\usepackage{amsmath,amsfonts} % hyperlinks
\usepackage{url}            % simple URL typesetting
\usepackage{booktabs}       % professional-quality tables
\usepackage{amsfonts}       % blackboard math symbols
\usepackage{pifont}
\usepackage{nicefrac}       % compact symbols for 1/2, etc.
\usepackage{microtype}      % microtypography
\usepackage{newtxtext}
\usepackage[font=small]{caption}
\usepackage{graphicx}
\usepackage{hyperref}
\usepackage{svg}
\usepackage{url}
\usepackage{multirow}

\usepackage{adjustbox}
\usepackage{subcaption}
\usepackage{stmaryrd}
\usepackage{dsfont}
\usepackage{tikz} % for drawing circles
\usepackage{wrapfig}
\newcommand{\circled}[1]{%
    \tikz[baseline=(char.base)]{
        \node[shape=circle, fill=black, text=white, inner sep=0.3pt, minimum size=8pt] (char) {\small #1};
    }%
}

\usepackage{iclr2025_conference,times}

\title{RESOLVE: Relational Reasoning with Symbolic and Object-Level Features Using Vector Symbolic Processing}
\iclrfinalcopy
\author{Mohamed Mejri, Chandramouli Amarnath \& Abhijit Chatterjee \\
School of Electrical and Computer Engineering \\
Georgia Institute of Technology \\
}
\begin{document}

\maketitle

\begin{abstract}
Modern transformer-based encoder-decoder architectures struggle with reasoning tasks due to their inability to effectively extract relational information between input objects (data/tokens). Recent work introduced the \textit{Abstractor} module, embedded between transformer layers, to address this gap. However, the Abstractor layer while excelling at capturing relational information (pure relational reasoning), faces challenges in tasks that require both object and relational-level reasoning (partial relational reasoning). To address this, we propose \texttt{RESOLVE}, a neuro-vector symbolic architecture that combines object-level features with relational representations in high-dimensional spaces, using fast and efficient operations such as bundling (summation) and binding (Hadamard product) allowing both object-level features and relational representations to coexist within the same structure without interfering with one another. \texttt{RESOLVE} is driven by a novel attention mechanism that operates in a bipolar high dimensional space, allowing fast attention score computation compared to the state-of-the-art. By leveraging this design, the model achieves both low compute latency and memory efficiency. \texttt{RESOLVE} also  offers better generalizability while achieving higher accuracy in purely relational reasoning tasks such as sorting as well as partial relational reasoning tasks such as math problem-solving compared to state-of-the-art methods.
\end{abstract}

\section{Introduction}
% Problem broad statement
\begin{wrapfigure}{R}{0.5\textwidth}
	\vskip-5pt
	\begin{tabular}{c}
		\includegraphics[width=.4\textwidth]{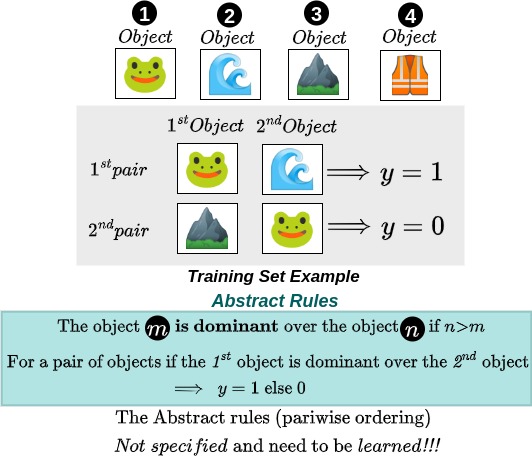}
	\end{tabular}
	\caption{\centering Example of purely relational task: \textit{Pairwise Ordering}}
 \label{pure}
 \vspace{-0.5cm}
\end{wrapfigure}
Analogical reasoning, which involves recognizing abstract relationships between objects, is fundamental to human abstraction and thought. This contrasts with semantic (meaning-based) and procedural (task-based) knowledge acquired from sensory information, which is typically processed through contemporary approaches like deep neural networks (DNNs). However, most of these techniques fail to extract abstract rules from limited samples~\cite{barrett2018measuring,ricci2018same,lake2018generalization}. 

\begin{figure}[htbp]
    % Second and third images - bottom row
    \begin{subfigure}[b]{0.38\textwidth} % 45% width for the 2nd figure
        \centering
        \includegraphics[width=0.8\textwidth]{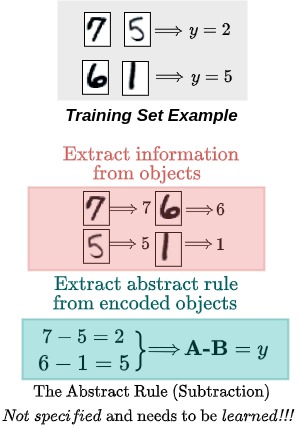}
        \caption{\centering  \textit{Subtraction using MNIST digits}}
        \label{partial_1}
    \end{subfigure}
    \hfill
    \begin{subfigure}[b]{0.38\textwidth} 
        \centering
        \includegraphics[width=1.2\textwidth]{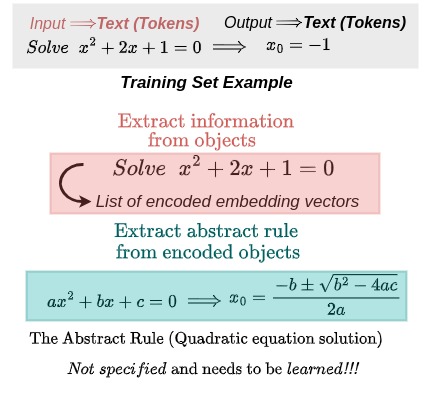}
        \caption{\centering  \textit{Quadratic Equation Solution}}
        \label{partial_2}
    \end{subfigure}
    \caption{\centering Two examples of partially relational tasks(Figure~\ref{partial_1} and \ref{partial_2})}
    \vspace{-0.25cm}
\end{figure}

These reasoning tasks can be \textit{purely} or \textit{partially} relational. Figure~\ref{pure} presents an example of a purely relational task where the objects (e.g. frog, mountains) are \textit{randomly} generated. In this task, only the information representing relationships \textit{between} the objects is relevant, not the objects themselves. 
%This makes the information located only at the level of relational representation—dominance, in this case—between objects and not the objects themselves. 
% Relational bottleneck is a perfaitly tailored for purely relational tasks where the object level features are not relevant. 
By contrast, in Figure~\ref{partial_1} the purpose is to learn the abstract rule of subtraction, which is unknown to the model, from pairs of MNIST digits. This abstract rule relies on the \textit{relational representation} between the digits (derived from their relationship with one another, in this case their ordering) and the digits themselves (the values being subtracted), which are \textit{object} features. This is a \textit{partially} relational problem.
% However, this abstract rule relies on the relational representation between the two digits (e.g., the order they appear in) but also on the value of the digits, materialized inside the object features, making it a partially relational problem. 
Similarly, in Figure~\ref{partial_2}, the purpose is to learn the abstract rule of the quadratic formula (i.e. the solution to the quadratic problem shown at the bottom of Figure~\ref{partial_2}) from the object features (derived from the text tokens representing equation coefficients) and the relational representation (derived from the coefficient ordering).
%induce the abstract rule—quadratic equation solutions in this case—from the object features (the text tokens, in this case) but also from the relational representation inside the input text (e.g., the relation between quadratic polynomial coefficients and the exact solution).

% Define what the rules are, what are partial/pure relational problems, define based on figure.
%These relational abstract rules can be \textit{purely} or \textit{partially} relational. An example of \textit{purely relational} features is shown in Figure \ref{pure}, where the sorting task is agnostic to the individual objects being sorted and solely depends on the ordering relation between them (the \textit{abstract rule} being learned). \textit{Partially} relational tasks combine both inherent features of the input objects and the relations between them, as shown in Figure \ref{partial_1} with subtraction and Figure \ref{partial_2} for quadratic equations. In both cases, the input itself requires parsing and feature extraction and the relational features between the input objects are mathematical operators. Both the relational features (operators) and objects (numbers/coefficients) influence the \textit{abstract rule} being learned (subtraction or quadratic formulae, as in the figures).

These relational or partially relational tasks have been shown to be problematic for transformer-based architectures \citep{abstractor}, which encode both the object features and relational representations into the same structure. \citep{abstractor} instead created a learnable inductive bias derived from the transformer architecture for explicit relational reasoning. Although this solution is sufficient for purely relational tasks such as Figure \ref{pure}, it is less efficient for partially relational tasks such as Figures \ref{partial_1} and \ref{partial_2} where the object features and relational representations are both significant.

The poor ability of transformers to superpose relational representations and object-level features is due to the low dimensionality of their components, causing interference between object features and relational representations \citep{webb2024relational}. By contrast, vector symbolic architectures (VSA) have used high-dimensional spaces to superpose object features and relational representations with low interference~\cite{rahimi}. Transformer-based architectures are moreover known to be power-inefficient due to the attention score computation~\cite{debus2023reporting}. Vector symbolic architectures have been proven to be power-efficient~\cite{menet2024mimonets} with low memory overhead due to the low-bitwidth (bipolar) representation of high-dimensional vectors.
However, current VSA techniques require prior knowledge of abstract rules and a pre-engineered set of relations and objects (e.g., blue, triangle), making them unsuitable for sequence-to-sequence reasoning.

%Key contribution
These arguments motivate the design of \texttt{RESOLVE}, an innovative vector symbolic architecture allowing superposition of relational representations and object-level features in high dimensional spaces. 
% We leverage the Hadamard product, also known as \textit{binding}, for fast. 
Object-level features are encoded through a novel, fast, and efficient \textit{HD-attention} mechanism. 
The key contributions of this paper are:
\begin{itemize}
    \item We are the \textit{first} to propose a strategy for addressing the relational bottleneck problem (capturing relational information between data/objects rather than input data/object attributes or features from limited training data) using a \textit{vector symbolic architecture}. 
    %We leverage the properties of hyperdimensional dimensional computing to limit the interference between object features and relational information. 
    %We use the Hadamard product to generate explicit bindings in a high-dimensional space, 
    Our method captures relational representations of input objects in a hyperdimensional vector space, while maintaining object-level information and features in the \textit{same representation structure}, while  minimizing their interference with each other at the same time.  The method outperforms prior art in tasks that require both pure and partial relational reasoning.
    %This allows better generalizability moving away from the \textit{generalizability-superposition trade-off}~\cite{bowers2016some} that most neural networks suffer from.
    
    \item We implement a novel, fast and efficient attention mechanism that operates directly in a bipolar ($\{\! -1\!,1\!\}$) high-dimensional space. Vectors representing relationships between symbols are learned, eliminating the need for prior knowledge of abstract rules required by prior work~\cite{rahimi}.s

    \item Our system significantly reduces computational costs by simplifying attention score matrix multiplication to bipolar operations and relying on lightweight high-dimensional operations such as the Hadamard product (also known as \textit{binding}).
    %It also takes advantage of the interference-friendly property of high dimensional space, to reduce the number of objects (i.e., elements in a sequence) by blending their representation. This significantly reduces the computational expenses of the model.
    
\end{itemize}

%In Section so and so
In the following section we discuss related prior work, followed by an overview of our symbolic HD-attention mechanism in Section \ref{overview}. We then discuss our vector-symbolic hyperdimensional attention mechanism and contrast it to the relational bottleneck approach in Section \ref{relational_bottleneck_comparison}. The \texttt{RESOLVE} encoder and hypervector bundling is then discussed in Section \ref{resolve_bundling} and the full architecture in Section \ref{resolve_full}. We then present experimental validation in Section \ref{results}, followed by conclusions.
%In the following sections, we introduce the intuition behind and explain our novel \textit{HD-Attention} mechanism for hyperdimensional computing. We then discuss the proposed \textit{vector symbolic architecture} to address the relational bottleneck problem in high-dimensional spaces. We compare the performance of \texttt{RESOLVE} with the Abstractor \cite{abstractor}, a standard transformer architecture \cite{vaswani2017attention}, and other state-of-the-art methods on simple discriminative relational tasks using various synthetic datasets, to demonstrate the reasoning capabilities of the proposed approach. Finally, we apply \texttt{RESOLVE} to purely relational sequence-to-sequence tasks (e.g., Sequence Sorting) and partially relational tasks (e.g., mathematical problem-solving tasks \cite{saxton2019analysing}) to illustrate its potential in real-world applications.

%that allows for the superposition of object-level and abstract-level information at high dimensionality 

%This paper presents a vector-symbolic framework for 
%explained in Figure~\ref{ABOB_O}. 
%It allows object and abstract-level information mixture (superposition) in a high dimensional space through fast and well-defined operations (i.e., binding). 

\section{Related Work}
% Below can move up front
%Several studies \cite{barrett2018measuring,ricci2018same,lake2018generalization} have demonstrated that abstract reasoning and relational representations are notable weaknesses in modern neural networks. %Large language models can tackle symbolic reasoning tasks to a certain extent but require extensive training data, highlighting their inability to generalize from limited examples. Consequently, significant effort has recently been directed towards addressing this deficiency.

% Symbolic learning paragraph. Merge with neurosymbolic?
To address the problem of learning abstract rules, \textit{symbolic AI} architectures such as the Relation Network \citep{santoro2017simple} propose a model for pairwise relations by applying a Multilayer Perceptron (MLP) to concatenated object features. Another example, PrediNet \citep{shanahan2020explicitly}, utilizes predicate logic to represent relational features.
% Merge with symbolic AI
Symbolic AI approaches combined with neural networks were leveraged by \textit{neurosymbolic learning}~\cite{manhaeve2018deepproblog,badreddine2022logic,barbiero2023relational,xu2018semantic} to improve this rule learning, with optimizations such as logical loss functions ~\cite{xu2018semantic} and logical reasoning networks applied to the predictions of a neural network~\cite{manhaeve2018deepproblog}. However, these systems require prior knowledge of the abstract rules guiding a task. They also require pre-implementation of object attributes~\cite{rahimi} (e.g., red, blue, triangle, etc.). This approach is only feasible for simple tasks (e.g., Raven’s Progressive Matrices~\cite{raven1938raven}) and is not appropriate for complex sequence-based partially relational tasks such as the quadratic solution of Figure \ref{partial_2}.

% Transformer-based approaches
For sequence-based partially relational tasks such as the math problem-solving of Figures \ref{partial_1} and \ref{partial_2} an\textit{ encoder-decoder structure with transformers} has been used~\cite{saxton2019analysing}. However, transformers often fail to capture explicit relational representations.

% Relational bottleneck
A solution to the shortcomings of encoder-decoder approaches is proposed in \citep{webb2024relational}, using the \textit{relational bottleneck} concept. This aims to separate relational representations learned using a learnable inductive bias from object-level features learned using connectionist encoder-decoder transformer architectures or DNNs. Several models are based on this idea: \textit{CoRelNet}, introduced in \citep{kerg2022neural}, simplifies relational learning by modeling a similarity matrix. A recent study \citep{esbn} introduced an architecture inspired by Neural Turing Machines (NTM) \citep{graves2014neural}, which separates relational representations from object features. Building on this concept, the Abstractor \citep{abstractor} adapted Transformers \citep{vaswani2017attention} for abstract reasoning tasks by creating an 'abstractor'—a mechanism based on cross-attention applied to relational representations for sequence-to-sequence relational tasks. A model known as the Visual Object Centering Relational Abstract architecture (OCRA) \citep{webb2024systematic} maps visual objects to vector embeddings, followed by a transformer network for solving symbolic reasoning problems such as Raven’s Progressive Matrices \citep{raven1938raven}. A subsequent study \citep{mondal2024slot} combined and refined OCRA and the Abstractor to address similar challenges. However, these relational bottleneck structures still suffer from the drawback of intereference between the relational representations and object features in deep layers due to their lower feature dimensionality \citep{webb2024relational}.
%whereas recent work on vector-symbolic architectures in high dimensions can superpose object and relational features while avoiding this interference.

% ~\cite{webb2024relational} argues that transformers based structure are prone to interference due to their low dimensionality representation. In contrast,~\cite{rahimi,webb2024relational} confirm that vector symbolic architecture can superpose different atomic items while avoiding curse of compisitionality. 

% Rahimi introduces VSAs
However, recent work \citep{rahimi} has shown that \textit{Vector Symbolic Architectures (VSAs)}, a neuro-symbolic paradigm using high-dimensional vectors \citep{kanerva2009hyperdimensional} with a set of predefined operations (e.g., element-wise addition and multiplication), exhibit strong robustness to vector superposition as an alternative to the relational bottleneck. In addition, Hyperdimensional Computing (HDC) is recognized for its low computational overhead \citep{mejri2024novel,amrouch2022brain} compared to transformer-based approaches.  However, prior work on VSAs has relied on pre-engineered set of objects and relations, limiting their applicability to sequence-to-sequence reasoning tasks.
%Below is redundant
%In contrast, Transformers \cite{vaswani2017attention} are known to be power-intensive and time-consuming.

% Small key contribs paragraph
In contrast to prior research, this paper is the first to leverage VSAs to efficiently combine object-level information with relational information in high-dimensional spaces, taking advantage of the lower interference between object features and relational representations in high dimensions. We also propose the first efficient attention mechanism for high-dimensional vectors (\textit{HD-Attention}).

\section{Overview}\label{overview}

In this section we present an overview of prior architectures used to learn abstract rules, and use them to illustrate the unique features of our VSA-based architecture.

\begin{figure*}[htbp]
    \centering
    \begin{subfigure}[b]{0.29\textwidth} % Slightly less wide for 1st figure
        \centering 
        \includegraphics[width=0.8\textwidth]{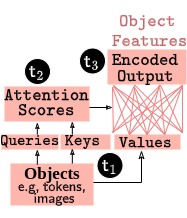}\vspace{0.5cm}
        \caption{{Self Attention} \\{(Transformer)}}
        \label{selfAttention_O}
    \end{subfigure}\hfill
    \begin{subfigure}[b]{0.29\textwidth} % Slightly less wide for 2nd figure
        \centering
        \includegraphics[width=\textwidth]{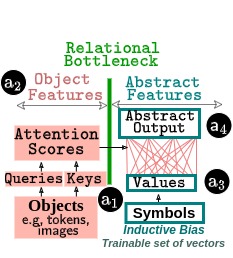}\vspace{0.4cm}
        \caption{Relational bottleneck \centering (Abstractor)}
        \label{abstractor_O}
    \end{subfigure}
    \hfill
    \begin{subfigure}[b]{0.36\textwidth} % Wider for the 3rd figure
        \centering
        \includegraphics[width=\textwidth]{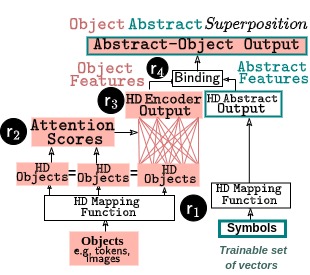}
        \caption{\centering Object and symbolic vector symbolic feature mixing (RESOLVE)}
        \label{ABOB_O}
    \end{subfigure}
    \small\caption{Comparison of a relational bottleneck approach applied on the transformer (Figure~\ref{abstractor_O}) separating object-level features while keeping only abstract features with a vector symbolic architecture alternative to the relational bottleneck using \textit{binding} to mix object and abstract level information in high dimensional (HD) space with low interference (Figure~\ref{ABOB_O})}
\end{figure*}

Figure~\ref{selfAttention_O} illustrates the self-attention mechanism used in transformer architectures~\cite{vaswani2017attention}. In step \circled{$\textsf{t}_1$}, objects are first encoded into keys, queries, and values. In step \circled{$\textsf{t}_{2}$}, self-attention captures correlations between keys and queries in an \textit{attention score} matrix. Finally, in step \circled{$\textsf{t}_{3}$}, this matrix is used to mix values and create encoded outputs. Self-attention is thus designed to capture correlations between input object sequence elements. However, it fails to capture relational representations of the input object sequence~\cite{abstractor}, leading to poor generalization capability for abstract rule-based tasks. The \textit{Abstractor} mechanism aims to fix that flaw.
% illustrates the abstractor mechanism~\cite{abstractor}.

Figure \ref{abstractor_O} illustrates the Abstractor mechanism. In step \circled{$\textsf{a}_1$} of the abstractor architecture shown in Fig. \ref{abstractor_O}, objects are first encoded into queries and keys, which are used to build attention scores (step \circled{$\textsf{a}_2$}) similar to self-attention. In parallel, in step \circled{$\textsf{a}_3$}, a set of symbols (learnable inductive biases) consisting of a set of trainable vectors are encoded into values. In step \circled{$\textsf{a}_4$}, the attention scores and the symbols are used to generate the abstract outputs, a dedicated structure for relational representations~\cite{abstractor,esbn} that are disentangled from object-level features. This approach, known as the \textit{relational bottleneck}, separates object-level features from relational representations. However, this separation can make it difficult to learn abstract rules for partially relational tasks.
\vspace{-0.1in}

%%%%%%%%%%%%%%%%%%%%%%%%%%%%%%%%%%%%%%%%%%%%%%%%%%%%%%%%%%%%%%%%%%%%%%%%%%%%%%%%%%%%%%%%%%%%%%%%%%%%
%%%%%%%%%%%%%%%%%%%%%%%%%%%%%%%%%%%%%%%%%%%%%%%%%%%%%%%%%%%%%%%%%%%%%%%%%%%%%%%%%%%%%%%%%%%%%%%%%%%%

%However, recent neuro-vector symbolic approaches that rely on high-dimensional computing either require a pre-engineered set of predicates and objects \cite{rahimi}, making them unsuitable for sequence-to-sequence tasks. 

% The curse of compositionality has also been addressed by neuro-symbolic approaches that use quasi-orthogonal high-dimensional vectors \cite{rahimi, menet2024mimonets} for storing relational representations. In \cite{menet2024mimonets}, it is shown that high-dimensional vectors are less prone to interference between object- and abstract features. 

% Rewrite the paragraph around the figure RESOLVE. !!!!!

The \texttt{RESOLVE} architecture (shown in Figure~\ref{ABOB_O}) explicitly structures the learning of relational information while encoding object-level features. In step \circled{$\textsf{r}_1$}, objects and symbols are mapped to a high-dimensional (HD) space using an high-dimensional encoder to generate HD Objects (object-level feature representations) and HD Abstract outputs (relational representations). The HD Objects, shown three times, are identical. They are first used in step \circled{$\textsf{r}_2$} to compute attention scores. Then, in step \circled{$\textsf{r}_3$}, these attention scores are used as weights to combine the HD Objects, producing an HD encoded output. In step \circled{$\textsf{r}_4$}, the HD Abstract output and the HD encoded output are superimposed through a binding operation (Hadamard product) to provide a mixed relational representation and object feature vector in high dimensions, avoiding the interference between relational representations and object features that this mixing causes in lower dimensions (seen in transformers \citep{webb2024relational}).

\section{Relational Bottleneck and VSA approach Modeling }\label{relational_bottleneck_comparison}
\begin{figure*}[htb]
    \centering
\vspace{-0.25cm}
    \begin{subfigure}[b]{0.44\textwidth}
        \centering
        \includegraphics[width=0.6\textwidth]{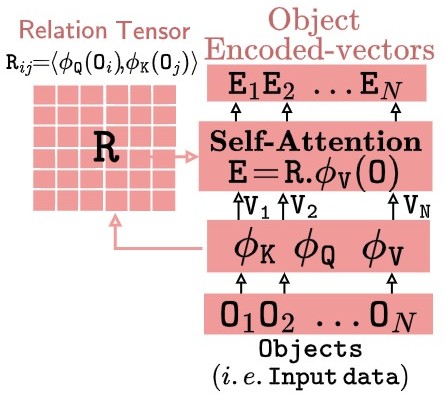} %\vspace{-0.25cm}
        \hspace{1cm}
        \caption{$\texttt{E} \gets \mathrm{SelfAttention}(\texttt{O})$}
        \label{fig:Self_attention}
    \end{subfigure}
    \hspace{-0.05\textwidth} % Adjusts the spacing between the subfigures
    \begin{subfigure}[b]{0.44\textwidth}
        \centering
        \includegraphics[width=0.6\textwidth]{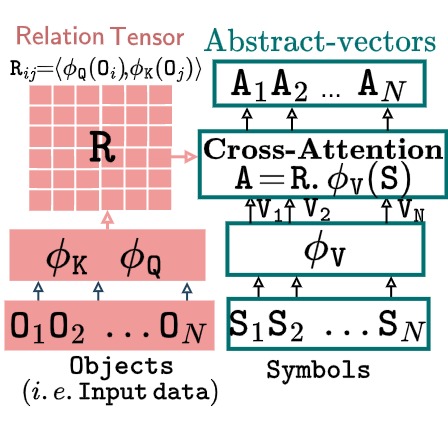} 
        \caption{$\texttt{A} \gets \mathrm{RelCrossAttention(\texttt{O}, \texttt{S})}$}
        \label{fig:Relational_cross_attention}
    \end{subfigure}
    
    \begin{subfigure}[b]{0.7\textwidth}
        \centering
        \vspace{0.1in}
        \includegraphics[width=0.7\textwidth]{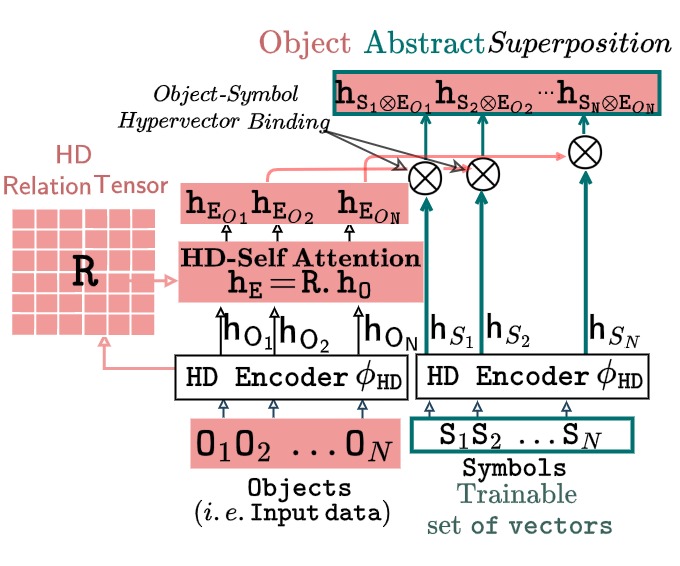} \caption{$\texttt{h}_{\texttt{E}_\texttt{O} \otimes \texttt{S}} \gets \mathrm{HD\text{-}Attention}(\texttt{h}_{\texttt{O}}) \otimes \texttt{h}_{\texttt{S}}$}
        \label{fig:VSA}
    \end{subfigure}    
    \caption{\footnotesize Comparison between \textrm{SelfAttention}~\cite{vaswani2017attention}~\ref{fig:Self_attention},  \textrm{RelationalCrossAttention}~\cite{abstractor}~\ref{fig:Relational_cross_attention} and our VSA approach~\ref{fig:VSA}. We show a single head of multi-attention for brevity. The object-related operations are in red and the relational-related (abstract/symbolic) operations are in turquoise. 
    %In red and turquoise respectively, the object-related and symbolic/abstract-based operations.     
    %the \texttt{objects} related by Softmax normalized relation tensor $\overline{R}(.,.)$; in green, the inductive bias denoted by \texttt{Symbols}.
    }
    \label{fig:self-vs-cross-attention}
    \vspace{-0.5cm}
\end{figure*}

Figure~\ref{fig:self-vs-cross-attention} contrasts the self-attention mechanism (Figure~\ref{fig:Self_attention}), the relational cross-attention (Figure~\ref{fig:Relational_cross_attention}), and our approach (Figure~\ref{fig:VSA}), using red and turquoise colors. The self-attention mechanism in Figure~\ref{fig:Self_attention} is applied to a sequence of objects (for instance, token embeddings) denoted by $\texttt{O}_\texttt{1..N}$. Each object is of dimension $\texttt{F}$. The objects are encoded into Keys, Queries, and Values through linear projections: $\phi_Q \colon \texttt{O} \mapsto \texttt{O} \cdot \texttt{W}_Q$, $\phi_K \colon \texttt{O} \mapsto \texttt{O} \cdot \texttt{W}_K$, and $\phi_V \colon \texttt{O} \mapsto \texttt{O} \cdot \texttt{W}_V$, where $\texttt{W}_Q,\texttt{W}_K,\texttt{W}_V$ are learnable matrices. The Queries and Keys are used to compute an attention score matrix, which captures the relationships between encoded objects through a pairwise dot product $\langle, \rangle$. \citep{abstractor} interprets this as a \textit{relation tensor}, denoted by $\texttt{R} = \left[\langle \phi_Q(\texttt{O}_i), \phi_K(\texttt{O}_j) \rangle\right]{i,j=1}^{N}$. $\texttt{R}$ is normalized to obtain $\overline{\texttt{R}}$ using a \texttt{Softmax} function to produce probabilities. $\mathrm{SelfAttention}(\texttt{O})$ thus generates a mixed relational representation $\texttt{E}$ of the encoded objects (Values) through the normalized relational tensor $\overline{\texttt{R}}$ (i.e., $\texttt{E}_i = {\sum}_{j};\overline{\texttt{R}}_{ij}\phi_V(\texttt{O}_j)$). A transformer uses the matrix $\overline{\texttt{R}}$ to capture input relations and $\phi_{V}$ to encode object-level features, but $\phi_{V}$ is not designed to learn abstract rules.

% \cite{abstractor} argues that transformers lack a dedicated structure for abstract rules and relational information, proposing a \textit{relational bottleneck} to separate object-level features from abstract rules. 
Figure~\ref{fig:Relational_cross_attention} shows the relational attention mechanism by \citep{abstractor} that isolates object-level features (in red) from abstract/relational information (in turquoise) to improve abstract rule learning. Like self-attention, the objects $\texttt{O}_{1..N}$ are first encoded into Keys and Queries through the same learnable projection functions $\phi_K$ and $\phi_Q$, which are then used to build a normalized relation tensor $\overline{\texttt{R}}$. In parallel, a set of symbols $\texttt{S}_{1..N}$ ($N$ learnable vectors with the same dimensionality as the objects) are encoded into values using a projection function $\phi_{V}$ (i.e., $V\!=\! \phi_{V}(\texttt{S})$). The encoded symbols (Values) are mixed using the relation tensor weights through a \textit{relational cross-attention} mechanism to generate a mixed relational representation containing less object-level information and more relational (abstract) information. These are called \textit{Abstract States}, denoted as $\texttt{A}_{1..N}$ ($\texttt{A}_{i} ={\sum}_{j} ;\overline{\texttt{R}}_{ij}\phi_V(\texttt{S}_j)$).

Figure~\ref{fig:VSA} shows our VSA-based system. It starts by encoding the objects $\texttt{O}_{1..N}$ from their $\texttt{F}$-dimensional space into a high-dimensional ($\texttt{D}$-dimensional) space using an encoder denoted by $\phi_\texttt{HD}$ to generate high-dimensional object vectors $\texttt{h}_{\texttt{O}_{1..N}}$. We extract relational scores from this using a novel \textit{HD-attention mechanism} to build a \textit{HD relation tensor}, denoted as $\texttt{R}$. This matrix is then normalized through a softmax function to generate $\overline{\texttt{R}}$. These normalized scores are used to mix the $\texttt{h}_{\texttt{O}_{1..N}}$, generating encoded object-level high-dimensional vectors $\texttt{h}_{\texttt{E}_{\texttt{O}_{i}}} = \sum_{j} \overline{\texttt{R}}_{ij} \texttt{h}_{\texttt{O}_{j}}$. A set of learnable symbols $\texttt{S}_{1..N}$ with the same dimensionality as the objects is used to encode relational information. These symbols are mapped to the high-dimensional space through $\phi_\texttt{HD}$, generating $h_{\texttt{S}_{1..N}}$. These high-dimensional symbolic vectors are \textit{bound} (i.e., Hadamard product/element-wise multiplication) with the encoded high-dimensional object-level vectors to generate vectors $\texttt{h}_{\texttt{E}_{\texttt{O} \otimes \texttt{S}}}$ that carry both object-level and relational (abstract) information.

%It starts by encoding the objects $\texttt{O}_{1..N}$ with a high dimensional encoder that maps objects from a low dimension $F$ to a high dimension $D$. The latter architecture inspired from~\cite{mejri2024adare} relies on $\mathcal{B} = \left[ \texttt{B}_\texttt{i} \right]_{\texttt{i}=0}^{\texttt{N}}$ $\in \mathbb{R}^{{N}\times{D-F+1}}$, a set $\texttt{N}$ of learnable high dimensional vectors called \textit{basis} and denoted $\texttt{B}$. Each object $\texttt{O}_\texttt{i}$ act as \textit{kernel} and is convolved with a \textit{basis} vector $\texttt{B}_\texttt{i}$ to generate an object level high dimensional vector $h_{\texttt{O}_{i}}$       

\section{RESOLVE: HD-Encoder and HD-Attention Mechanism With Hypervector Bundling}\label{resolve_bundling}
% \begin{figure*}[htpb]
%     \centering
%     \includegraphics[width=0.75\linewidth]{HD_Attention.jpg}
%     \caption{HD-Encoder $\phi_{\texttt{HD}}$ and HD-Attention($\texttt{O}_{\texttt{1\dots N}}$)}
%     \label{fig:hd_encode_attention}
%     \vspace{-0.25cm}
% \end{figure*}
\begin{wrapfigure}{R}{0.63\textwidth}
	\vskip-5pt
	\begin{tabular}{c}
		\includegraphics[width=.63\textwidth]{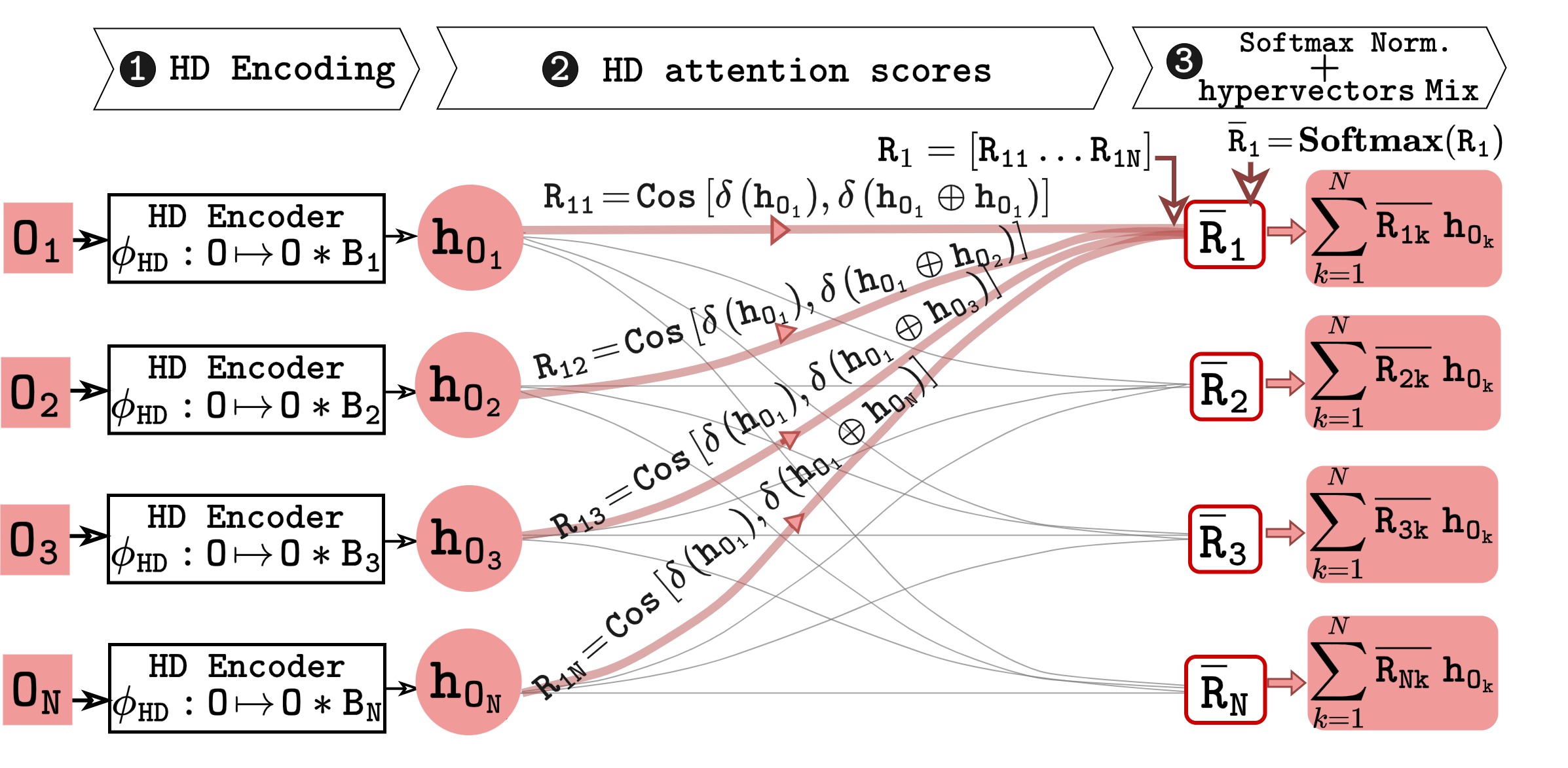}\\[-5pt]
	\end{tabular}
	\caption{\footnotesize HD-Encoder $\phi_{\texttt{HD}}$ and HD-Attention($\texttt{O}_{\texttt{1\dots N}}$)}
 \label{fig:hd_encode_attention}
\end{wrapfigure}

Figure~\ref{fig:hd_encode_attention} shows the HD-Encoder and HD-Attention mechanism applied to an input sequence of objects $\texttt{O}_{\texttt{1}}$, ..., $\texttt{O}_{\texttt{N}}$ (see Figure~\ref{fig:self-vs-cross-attention}). In Step \circled{\texttt{1}}, objects are mapped from the $F$-dimensional feature space to a $D$-dimensional HD space ($D\sim 10^3$) using the HD encoder $\phi_{\texttt{HD}}$. We have implemented $\phi_{\texttt{HD}}$ using single-dimension convolution operations inspired by ~\cite{mejri2024adare}. In this encoder scheme, each object $\texttt{O}_{\texttt{i}}$ is convolved with a learnable high dimensional vector called the \textit{HD-Basis} denoted by $\texttt{B}_{i}\in\mathbb{R}^{{N}\times{D-F+1}}$, giving rise to the high-dimensional HD Object hypervectors $\texttt{h}_{\texttt{O}_{\texttt{i}}}[j] = \sum_k \texttt{O}_{\texttt{i}}[k] \cdot \texttt{B}_{\texttt{i}}[j - k]$. 

Step \circled{\texttt{2}} consists of generating the relation tensor $\texttt{R}$, made up of attention scores that capture relationships between different HD objects $\texttt{h}_{\texttt{O}_{i}}$. These scores are built using a novel hyperdimensional attention mechanism, called \textit{HD-Attention}. %and are modeled by a pairwise interaction (i.e., \textit{inner product}) between encoded objects using learnable query and key encodings $\texttt{R}_{ij \in \llbracket 0, N \rrbracket}$ \cite{abstractor}.
Prior work~\cite{vaswani2017attention,abstractor} has generated these relational representations using a pairwise inner product between object features in a low dimensional space. In contrast, the HD-Attention mechanism maps object features to a high-dimensional space where (as shown in \cite{menet2024mimonets}), the HD Object hypervectors are quasi-orthogonal, allowing efficient relational representation and object feature superposition in the high dimensional vector space.
%In contrast, the object hypervectors are mapped into high D-dimensional space with more than 1000 elements per vector. \cite{menet2024mimonets} shows that in high-dimensional space hypervectors are quasi-orthogonal meaning that their cosine similarity is almost zero.  

%In prior work, the query and key embeddings are represented in a low-dimensional space (i.e., dimension less than one hundred). In this research, objects are projected onto a \textit{high D-dimensional hyperspace} (i.e., $D \geq 1000$) with data represented as \textit{hypervectors}. As shown in \cite{menet2024mimonets}, as the dimensionality of bipolar hypervectors increases, they become more \textit{orthogonal}, making This makes applying relational modeling approaches such as \cite{abstractor} difficult in hyperdimensional space. Using the inner product on \textit{orthogonal} vectors results in very low attention scores that fail to capture actual object correlations and relationships.

% One way to address this issue is to model the relationship between objects \textit{indirectly} by measuring their correlation to their superposition. To illustrate this, consider two objects, \(A\) and \(B\). Instead of directly computing the correlation between \(A\) and \(B\), we assess the relationship between object \(A\) and a vector formed by the sequence "\(AB\)."

The HD-Attention mechanism represents object sequences using the \textit{bundling} operation (i.e., \(\oplus\)) between HD-encoded sequence elements. Given two objects \(\texttt{O}_\texttt{i}\) and \(\texttt{O}_\texttt{j}\) we first project them onto a hyperspace using the \textit{HD-Encoder} $\phi_{\texttt{HD}}$. The HD object hypervectors are thus $\mathrm{h}_{\texttt{O}_i} = \phi_{\texttt{HD}}(\texttt{O}_\texttt{i})$. Before calculating the attention score, these HD objects are made bipolar using the function $\delta(x) = -\mathds{1}_{\{x < 0\}} + \mathds{1}_{\{x > 0\}}$, replacing the binary coordinate-wise majority in the bipolar domain used in \citep{sign}.
Thus, the $(i,j)$th element of the relation tensor \(\texttt{R}_{ij}\) denoting the object-level relationships between $\texttt{O}_\texttt{i}$ and $\texttt{O}_\texttt{j}$ can be expressed according to the equation~\ref{eq:hd_attention} where $cos(.)$ denotes the cosine similarity function and $\lVert \rVert_2$ denotes the L2 norm:
\begin{equation}
    \vspace{-0.1in}
     \texttt{R}_{ij} = \cos(\delta(\mathrm{h}_{\texttt{O}_i}), \delta(\mathrm{h}_{\texttt{O}_i} \oplus \mathrm{h}_{\texttt{O}_j})) = \frac{\langle \delta(\mathrm{h}_{\texttt{O}_i}), \delta(\mathrm{h}_{\texttt{O}_i} \oplus \mathrm{h}_{\texttt{O}_j}) \rangle}{D}
    \label{eq:hd_attention}
\end{equation}

The denominator of the cosine similarity function $cos(.)$ is ${\lVert \delta(\mathrm{h}_{\texttt{O}_i} )\rVert}_{2}.{\lVert \delta(\mathrm{h}_{\texttt{O}_i}\!\oplus\!\mathrm{h}_{\texttt{O}_j} )\rVert}_{2}$. Since the HD objects are bipolar, their L2 norm is $\sqrt{D}$, leading to the expression in Equation~\ref{eq:hd_attention}.  
We define \textit{bundling} ($\oplus$) as the element-wise real value summation between two bundled HD objects.
% The bundling operation implicitly acts as a majority vote between two bipolar object elements. 
It captures the \textit{dominant} or \textit{relevant} features of an object pair. The sign of each HD object element follows the sign of the element with a higher magnitude, amplified by dominant features of object pair during training.
In step, \circled{\texttt{3}} the relation tensor matrix $\texttt{R}$ is normalized using a softmax function to generate $\overline{\texttt{R}}$. This matrix is used to encode the HD Object hypervectors by mixing them according to their corresponding weights in the normalized relation tensor $\overline{\texttt{R}}$.

\section{RESOLVE: Architecture Overview}\label{resolve_full}

\begin{figure*}[htb]
    \centering    \includegraphics[width=0.8\textwidth]{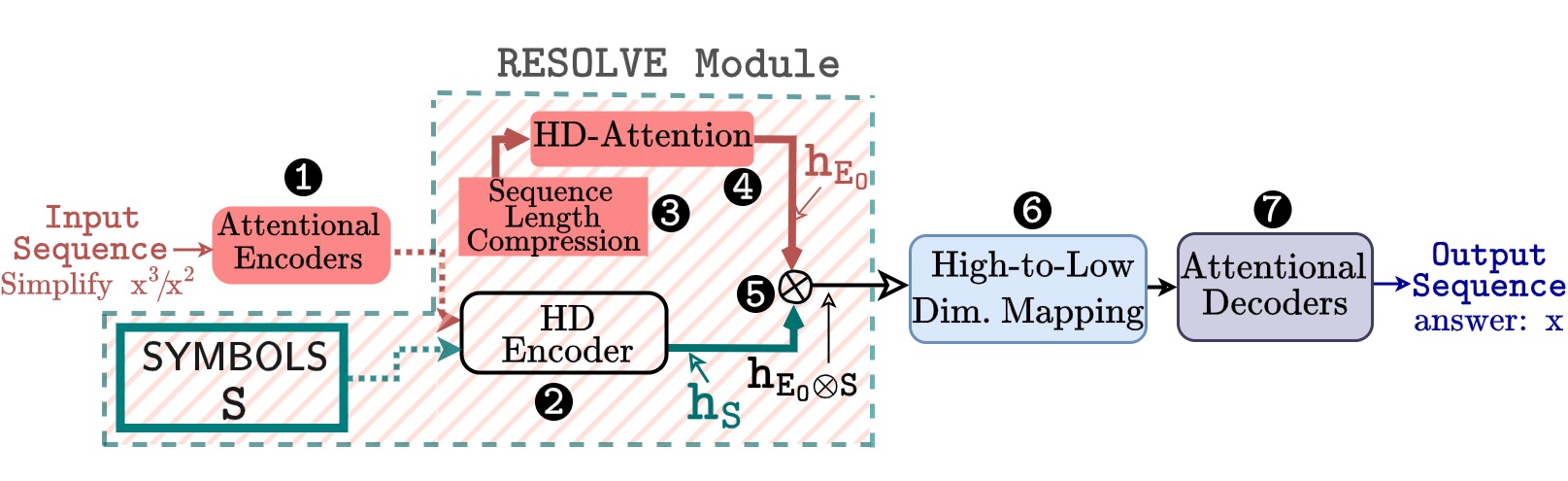}
    \caption{\small The \texttt{RESOLVE} module inside an encoder-decoder for sequence-to-sequence tasks. It includes abstract (turquoise path) and object level (red path) information that are superposed. The \texttt{RESOLVE} module operates in a high dimensional space (bold arrows). The rest operate in a low dimensional space (dotted arrows)}
    \label{fig:AB-OB-Overview}
    \vspace{-0.52cm}
\end{figure*}
\begin{wrapfigure}{R}{0.43\textwidth}
    \vspace{-0.1in}
    \centering
    \begin{subfigure}{0.3\textwidth}
        \centering
        \includegraphics[width=\textwidth]{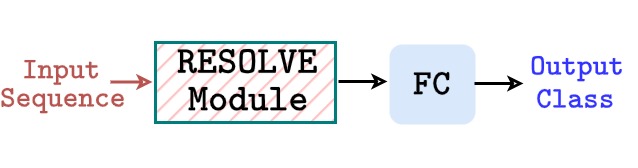}
        \captionsetup{font=scriptsize}
        \caption{\texttt{RESOLVE} Architecture for \textit{single output purely relational task}}
        \label{fig:sub1-a}
    \end{subfigure}
    \hfill
    \begin{subfigure}{0.4\textwidth}
        \centering
        \includegraphics[width=\textwidth]{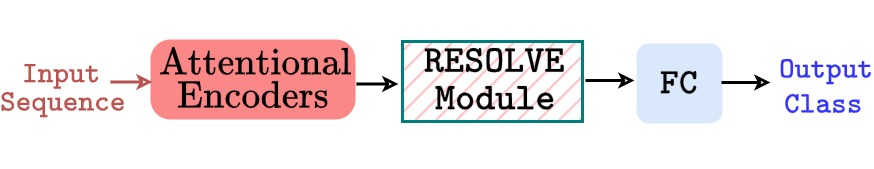}
        \captionsetup{font=scriptsize}
        \caption{\texttt{RESOLVE} Architecture for \textit{single output partially relational task}}
        \label{fig:sub1-b}
    \end{subfigure}
    \hfill
    
    \begin{subfigure}{0.45\textwidth}
        \centering
        \includegraphics[width=\textwidth]{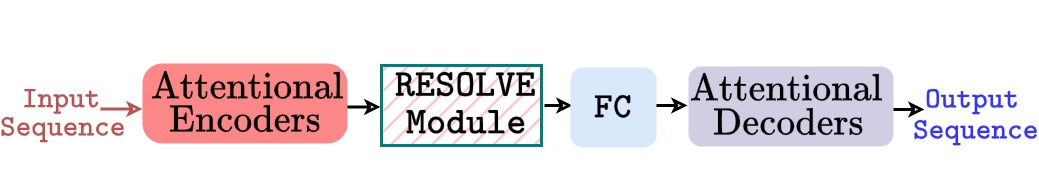}
        \captionsetup{font=scriptsize}
        \caption{\texttt{RESOLVE} Architecture for sequence-to- sequence \textit{purely relational task}}
        \label{fig:sub2}
    \end{subfigure}
    \hfill
    \begin{subfigure}{0.45\textwidth}
        \centering
        \includegraphics[width=\textwidth]{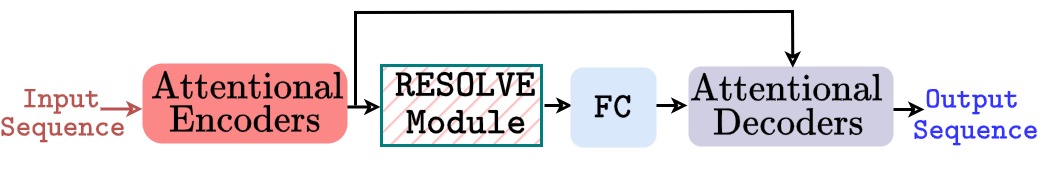}
        \captionsetup{font=scriptsize}
        \caption{\texttt{RESOLVE} Architecture for sequence-to- sequence \textit{partially relational task}}
        \label{fig:sub3}
    \end{subfigure}
    \caption{\small \texttt{RESOLVE} pipelines for four tasks}
    \label{fig:architecture}
    \vspace{-0.5cm}
\end{wrapfigure}
The \texttt{RESOLVE} module implementation is illustrated in Figure~\ref{fig:AB-OB-Overview}, for a sequence-to-sequence encoder-decoder structure with \texttt{RESOLVE} Modules (\circled{\texttt{2}} to \circled{\texttt{4}}). An input sequence, in this case a set of tokens, is encoded into embedding vectors and then passed to the \textrm{Attentional Encoders} in Step \circled{\texttt{1}}, which consist of self-attention layers followed by feedforward networks~\cite{vaswani2017attention}. This module is commonly used in sequence-to-sequence modeling and in prior art \citep{abstractor} to extract object-level information from the sequence.

In step \circled{\texttt{2}}, the output of the \textrm{Attentional Encoders}, which consists of a set of encoded objects, is mapped to a high-dimensional space using the \textrm{HD Encoder} \circled{\texttt{2}}. 
These HD Object hypervectors are then mixed using the \textrm{HD-Attention} \circled{\texttt{3}} mechanism to generate $\texttt{h}_{\texttt{E}_\texttt{O}}$.
In parallel, a set of relational representations (the learnable symbols of Section \ref{relational_bottleneck_comparison}) $\texttt{S}$ are mapped to high-dimensional space through the same HD-encoder \circled{\texttt{2}}. The resulting hypervectors, denoted by $h_s$, are then combined with the mixed HD Object hypervectors through a \textit{binding} operation in Step \circled{\texttt{4}}.
The result is denoted as $\texttt{h}_{{S \otimes \texttt{E}_\texttt{O}}}$.

High dimensional vectors are known to be holistic~\cite{kanerva2009hyperdimensional} meaning that information is uniformly distributed across them. This gives high information redundancy and makes it possible to map to a low-dimensional space with low information loss~\cite{yan2023efficient}. The hypervectors gained from Step \circled{\texttt{4}} ($\texttt{h}_{{S \otimes \texttt{E}_\texttt{O}}}$) are thus mapped to low-dimensional space through a learnable linear layer in Step \circled{\texttt{5}}. The resulting vectors are then forwarded to a set of \textrm{Attentional Decoders} in Step \circled{\texttt{6}}, which consist of causal-attention and cross-attention layers~\cite{vaswani2017attention}. 
% The regular attention mechanism in the decoder~\cite{vaswani2017attention} justifies the need for high-to-low dimensionality mapping as they rely on time-consuming matrix multiplications, which become increasingly inefficient in high dimensional space.

Figure~\ref{fig:architecture} shows four different \texttt{RESOLVE} architecture configurations used for different tasks. 
The \texttt{RESOLVE} architectures illustrated in (Figure~\ref{fig:sub1-a}) consists of a single \texttt{RESOLVE} encoder followed by a fully connected layer. It is used for single output purely relational tasks (e.g. pairwise ordering) that don't require an attentional object level encoding. On the other hand, Figure~\ref{fig:sub1-b} shows the same architecture with an attentional encoder in the front-end used to process object level features for single output partially relational tasks (e.g. learning the abstract rule of subtraction).  
Figure~\ref{fig:sub2} and \ref{fig:sub3} shows \texttt{RESOLVE} architecture for sequence-to-sequence purely (e.g. sorting) and partially relational tasks (e.g. mathematical problem solving \citep{saxton2019analysing}) respectively. Both of them use an attentional encoder to process object level features and an attentional decoder to generate the output sequence. However, the architecture in Figure~\ref{fig:sub3} requires a skip-connection between the encoder and the decoder because the output sequence in the partially relational tasks relies on object features as well as relational representations. 
\section{Experiments}\label{results}
We have evaluated the performance of \texttt{RESOLVE} compared to the state-of-the-art on several relational tasks: (1) Single output purely relational tasks (pairwise ordering, a sequence of image pattern learning with preprocessed inputs); (2) Single output partially relational tasks (sequence of image pattern learning with low-processed inputs, mathematical abstract rule learning from images); (3) Sequence-to-sequence purely relational tasks (Sorting); (4) Sequence-to-sequence partially relational tasks (Mathematical problem solving). The baselines used for comparison are CorelNet~\cite{kerg2022neural} with Softmax activation, Predinet~\cite{shanahan2020explicitly}, the Abstractor~\cite{abstractor}, the transformer~\cite{vaswani2017attention}, a multi-layer-perceptron (as evaluated in~\cite{abstractor}) and the LEN~\cite{zheng2019abstract}, a neuro symbolic architecture.            
\subsection{Single Output Purely Relational Tasks}
\vspace{-0.15in}
\begin{figure}[ht]
    %\vskip-.2in
    \centering
    \includegraphics[width=0.7\textwidth]{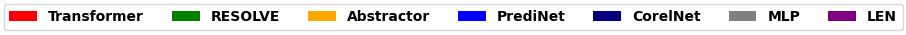}
    \begin{subfigure}[t]{0.43\textwidth}
        \centering\captionsetup{width=.9\linewidth}
        % \vskip-20pt
        \includegraphics[width=\textwidth]{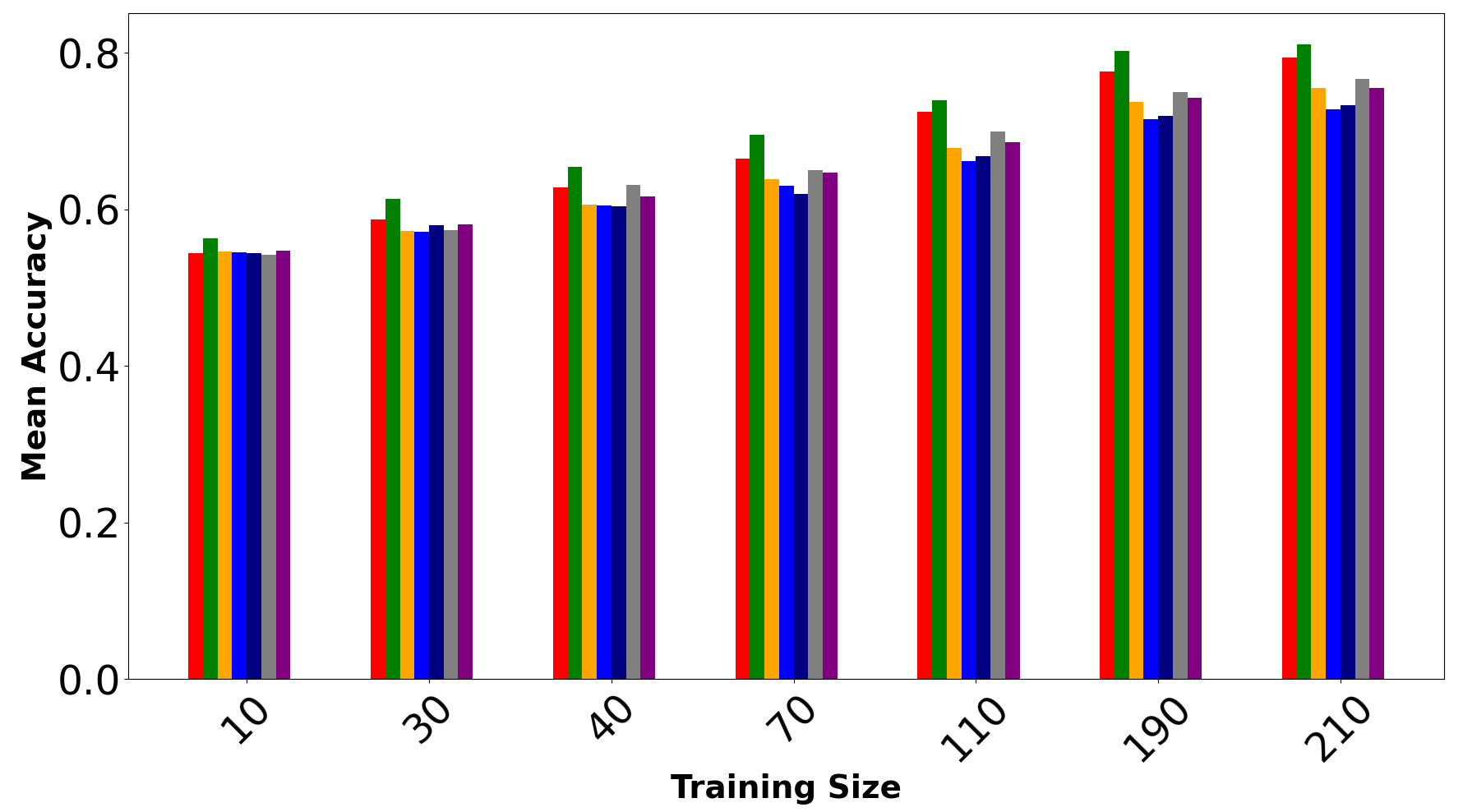}
        \vskip-5pt
        \caption{Learning $\prec$ relation  accuracy}\label{fig:exp_order_relation}
    \end{subfigure}
    % \hskip10pt
    % \captionsetup[subfigure]{oneside,margin={-.3in,0in}}
    \begin{subfigure}[t]{0.41\textwidth}
        \centering\captionsetup{width=.9\linewidth}
        % \vskip-20pt
        \includegraphics[width=\textwidth]{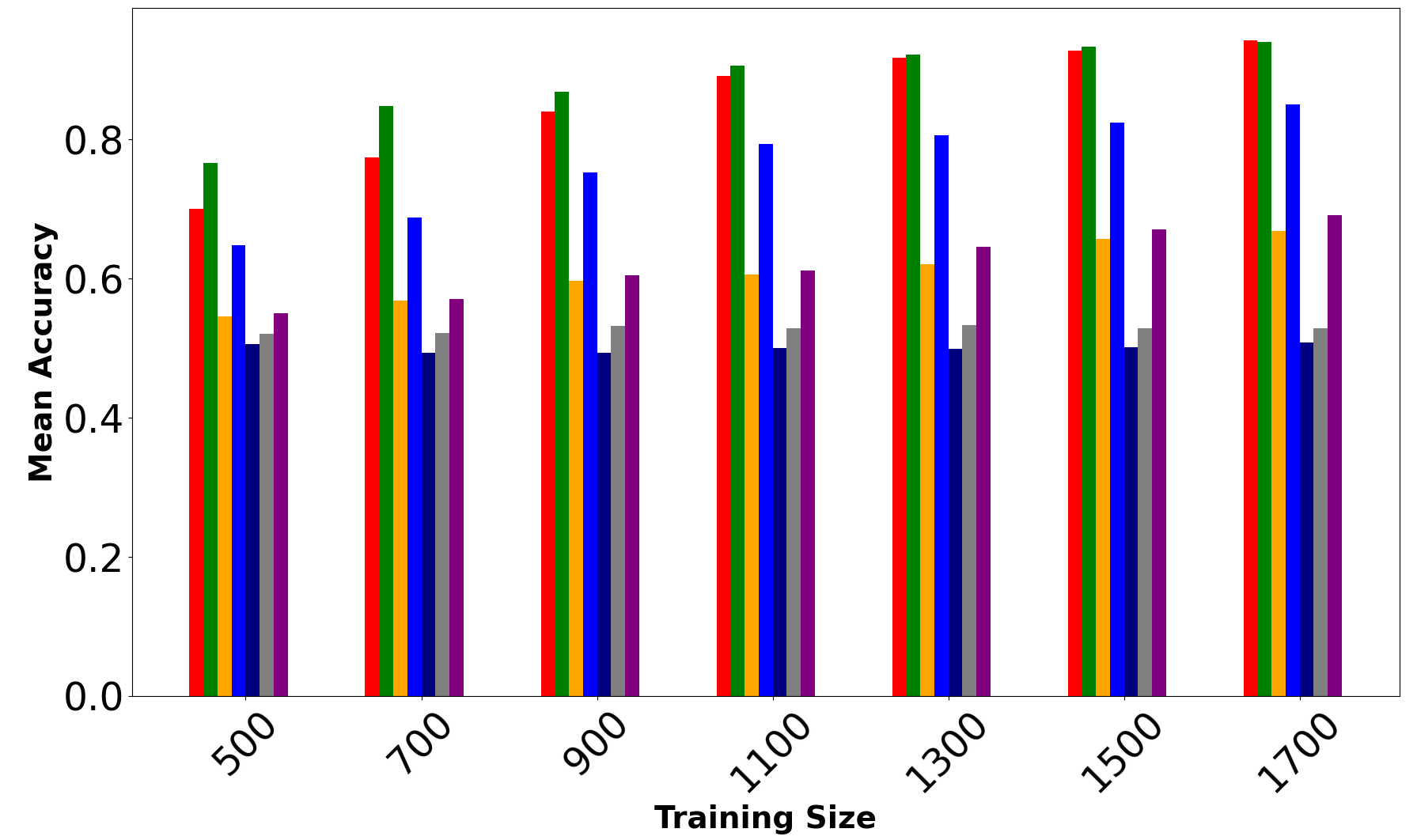}
        \vskip-5pt
        \caption{SET Classification accuracy}\label{fig:exp_set_classification}
    \end{subfigure}
    \caption{Experiments on single output \textit{purely }relational tasks and comparison to SOTA.}
    \vskip-20pt
\end{figure}

\paragraph{Order relations: modeling asymmetric relations} As described in ~\cite{abstractor}, we generated 64 random objects represented by iid Gaussian vectors \( o_i \sim \mathcal{N}(0, I) \in \mathbb{R}^{32} \), and established an anti-symmetric order relation between them \( o_1 \prec o_2 \prec \cdots \prec o_{64} \). From 4096 possible object pairs \((o_i, o_j)\), 15\% are used as a validation set and 35\% as a test set. We train models on varying proportions of the remaining 50\% and evaluate accuracy on the test set, conducting 5 trials for each training set size. The models must generalize based on the transitivity of the \( \prec \) relation using a limited number of training examples. The training sample sizes range between 10 and 210 samples. Figure \ref{fig:exp_order_relation} demonstrates the high capability of \texttt{RESOLVE} to generalize with just a few examples, achieving over $80\%$ accuracy with just 210 samples ($1.05 \times$ better than the second best model and $1.09 \times$ better than Abstractor). The Transformer model is the second best performer, better than the Abstractor and CorelNet-Softmax due to the lower level of abstraction needed for learning asymmetric relations.

\paragraph{\textit{SET}: modeling multi-dimensional relations with \textit{pre}-processed objects}\label{set-parag}
\begin{wrapfigure}{R}{0.2\textwidth}
	\vskip-5pt
	\begin{tabular}{c}
		\includegraphics[width=.2\textwidth]{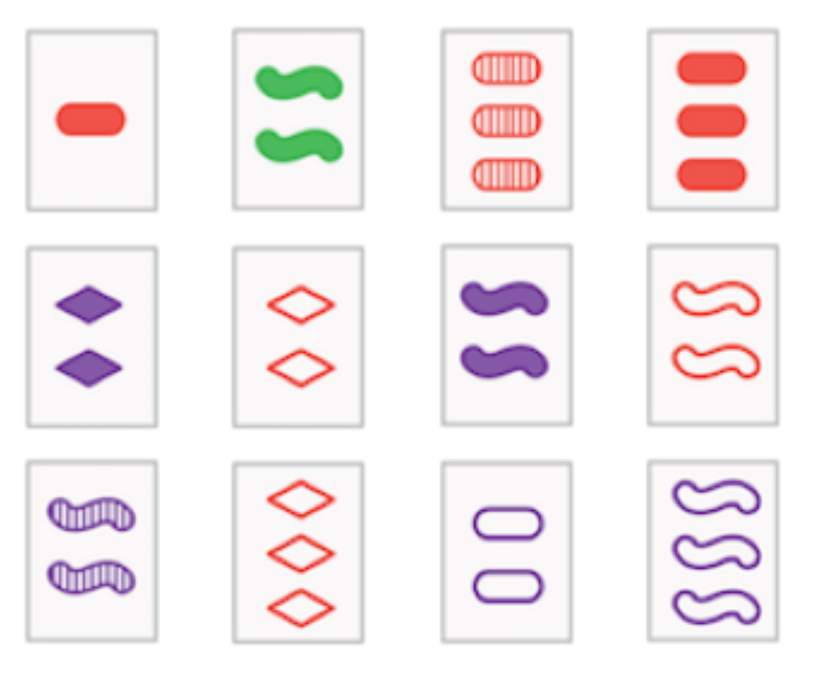}\\[-5pt]
	\end{tabular}
	\caption{\footnotesize The SET game}
 \label{set_figure}
 \vspace{-0.5cm}
\end{wrapfigure}
In the SET \citep{abstractor} task, players are presented with a sequence of cards. Each card varies along four dimensions: color, number, pattern and shape. A triplet of cards forms a "set" if they either all share the same value or each have a unique value (as in Figure \ref{set_figure}). The task is to classify triplets of card images as either a "set" or not. The shared architecture for processing the card images in all baselines as well as \texttt{RESOLVE} is $\texttt{CNN} \to \{\cdot\} \to \texttt{Flatten} \to \texttt{Dense}$, where $\{\cdot\}$ is one of the aforementioned modules. The CNN embedder is pre-trained and object features are taken from the last linear layer of the model. The relational models thus focus on learning the abstract rules without having to encode object features. For this specific task, there are four relational representations (e.g., shape, color, etc.) and one abstract rule (whether it is a triplet or not).

Figure \ref{fig:exp_set_classification} shows \texttt{RESOLVE} outperforms all the baselines (up to $1.05$x better than the second best model and $1.11$x better than the Abstractor), as it balances object features with relational representations. In this particular case, PrediNet also shows high accuracy. Its feature vectors are less connected to object-level features than those of the transformer but more than those of the Abstractor. This experiment shows that for descriminative purely relational tasks, abstract rules are often easy to extract and are highly correlated to object features, resulting in the transformer outperforming the Abstractor.  

\subsection{Single Output Partially Relational Tasks}
\begin{figure}[ht]
    %\vskip-.2in
    \centering
    \includegraphics[width=0.8\textwidth]{legend.png}
    \begin{subfigure}[t]{0.45\textwidth}
        \centering\captionsetup{width=.9\linewidth}
        % \vskip-20pt
        \includegraphics[width=\textwidth]{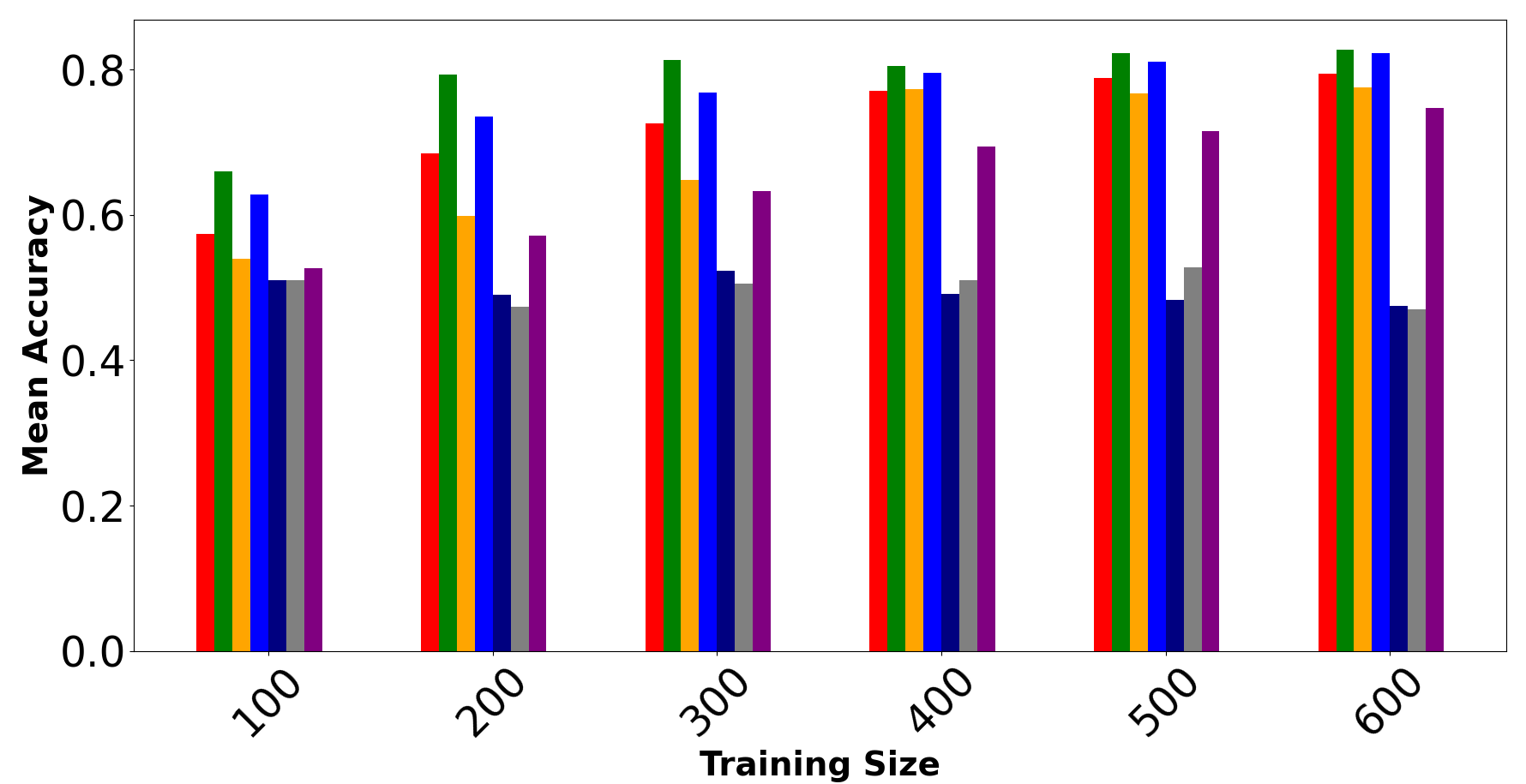}
        \vskip-5pt
        \caption{SET Classification accuracy}\label{fig:exp_set_classification_l}
    \end{subfigure}    
    \begin{subfigure}[t]{0.42\textwidth}
        \centering\captionsetup{width=.9\linewidth}
        % \vskip-20pt
        \includegraphics[width=\textwidth]{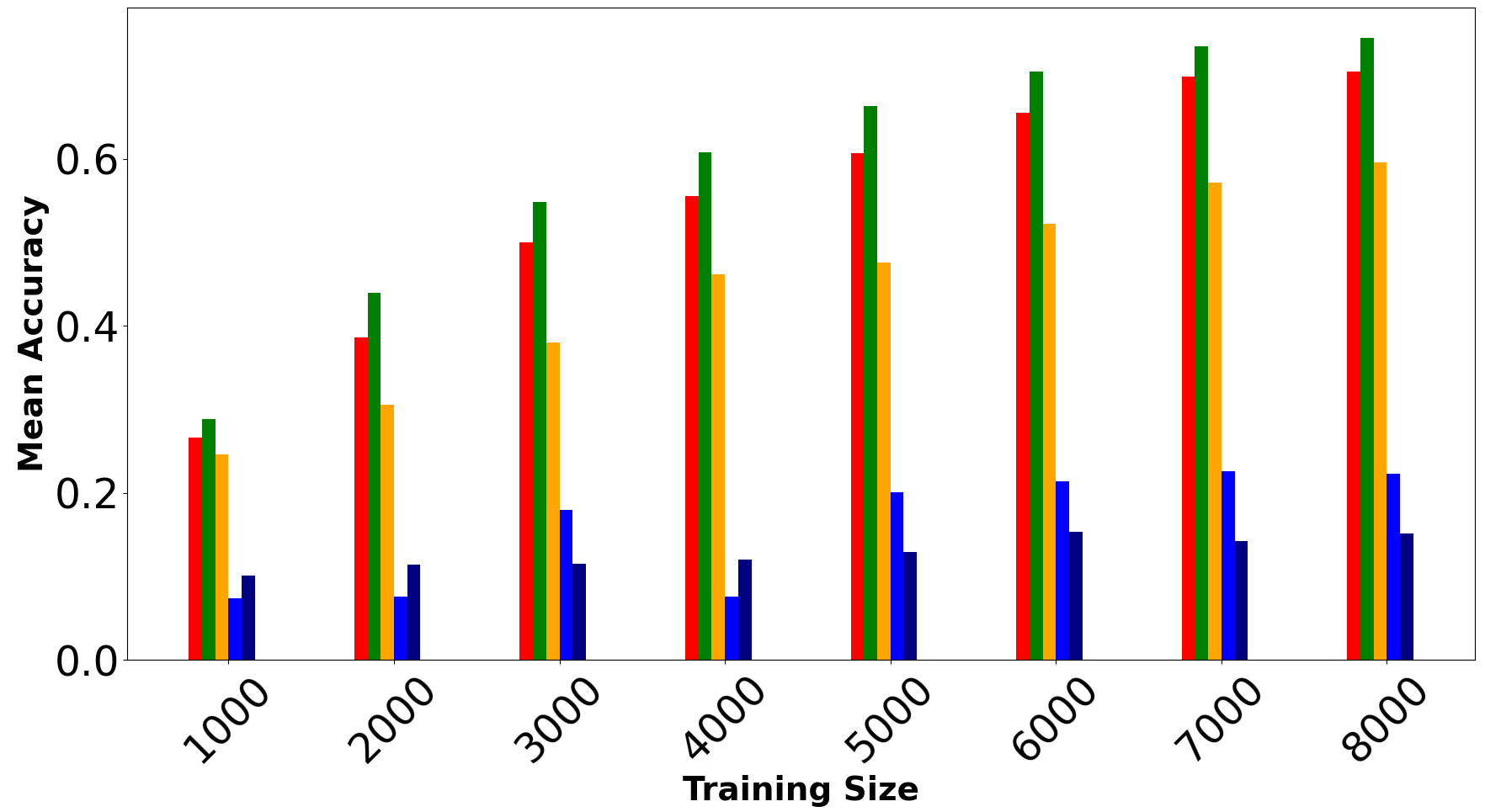}
        \vskip-5pt
        \caption{MNIST-Math Classification accuracy}\label{fig:result_mnist}
    \end{subfigure}
    % \hskip10pt
    % \captionsetup[subfigure]{oneside,margin={-.3in,0in}}
\vspace{-0.3cm}
    \caption{Experiments on single output \textit{partially} relational tasks and comparison to SOTA.}
    \vspace{-0.7cm}
\end{figure}
\vspace{-0.05in}

\paragraph{\textit{SET}: modeling multi-dimensional relations with \textit{low}-processed objects}
Instead of extracting highly encoded object level features from the pre-trained \texttt{CNN} used in Section~\ref{set-parag}, we extract the feature map of the first convolutional layer of the pretrained \texttt{CNN} to assess the ability of \texttt{RESOLVE} to handle low processed object level features. Figure~\ref{fig:exp_set_classification_l} shows the mean accuracy of different relational models when trained on small portion of the dataset. \texttt{RESOLVE} outperforms the state of the art with more than $80\%$ accuracy using just 600 training samples. In contrast to the Section~\ref{set-parag}, PrediNet is the second best model thanks to its balanced trade-off between object-level feature processing and abstract feature encoding. 
%\texttt{RESOLVE} is up to $1.08 \times$ better than PrediNet and $1.32 \times$ better than the Abstractor. 
\vspace{-0.1in}

\paragraph{\textit{MNIST-MATH}: extracting mathematical rules from a pair of digit images}
\begin{wrapfigure}{R}{0.42\textwidth}
	\vskip-5pt
	\begin{tabular}{c}
		\includegraphics[width=.42\textwidth]{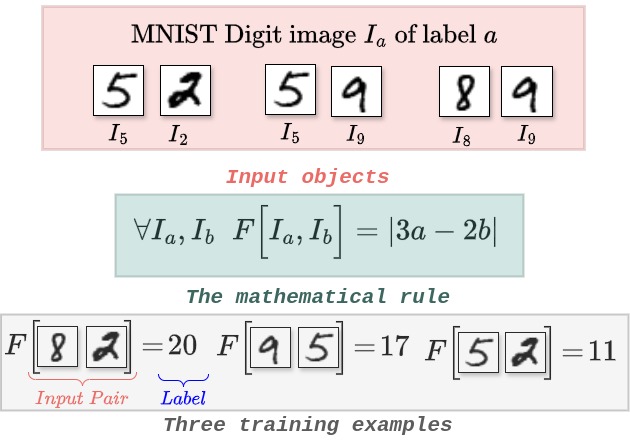}\\[-5pt]
	\end{tabular}
	\caption{\footnotesize MNIST-Math classification task}
 \label{fig:MNIST_MATH_last}
 \vspace{-0.8cm} 
\end{wrapfigure}
In this case (Figure~\ref{fig:MNIST_MATH_last}), the math rule to extract is a non-linear weighted subtraction (i.e, $F(a,b) = \lvert 3a-2b\rvert$). This task is partially relational since the input image label is unknown, making object level feature extraction more critical. 
%We should note that this is a classification problem with 29 classess corresponding to $\text{Card}\big(F(a,b), \{a,b\} \in \llbracket 0,9\rrbracket^2\big)$.  
According to Figure~\ref{fig:result_mnist}, \texttt{RESOLVE} outperforms the other baselines, with $1.14 \times$ better accuracy than the transformer and $1.47\times$ better accuracy than the Abstractor. The transformer outperforms the Abstractor here due to to the relative simplicity of the abstract rule, this problem relies more on object level information than abstract information.

\subsection{Object-sorting: Purely Relational Sequence-to-Sequence Tasks}\label{object_sorting}
\begin{wrapfigure}{R}{0.45\textwidth}
	\vskip-5pt
	\centering
	\includegraphics[width=.45\textwidth]{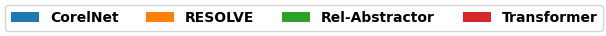}\\[-5pt] % Include the legend at the top
	\includegraphics[width=.45\textwidth]{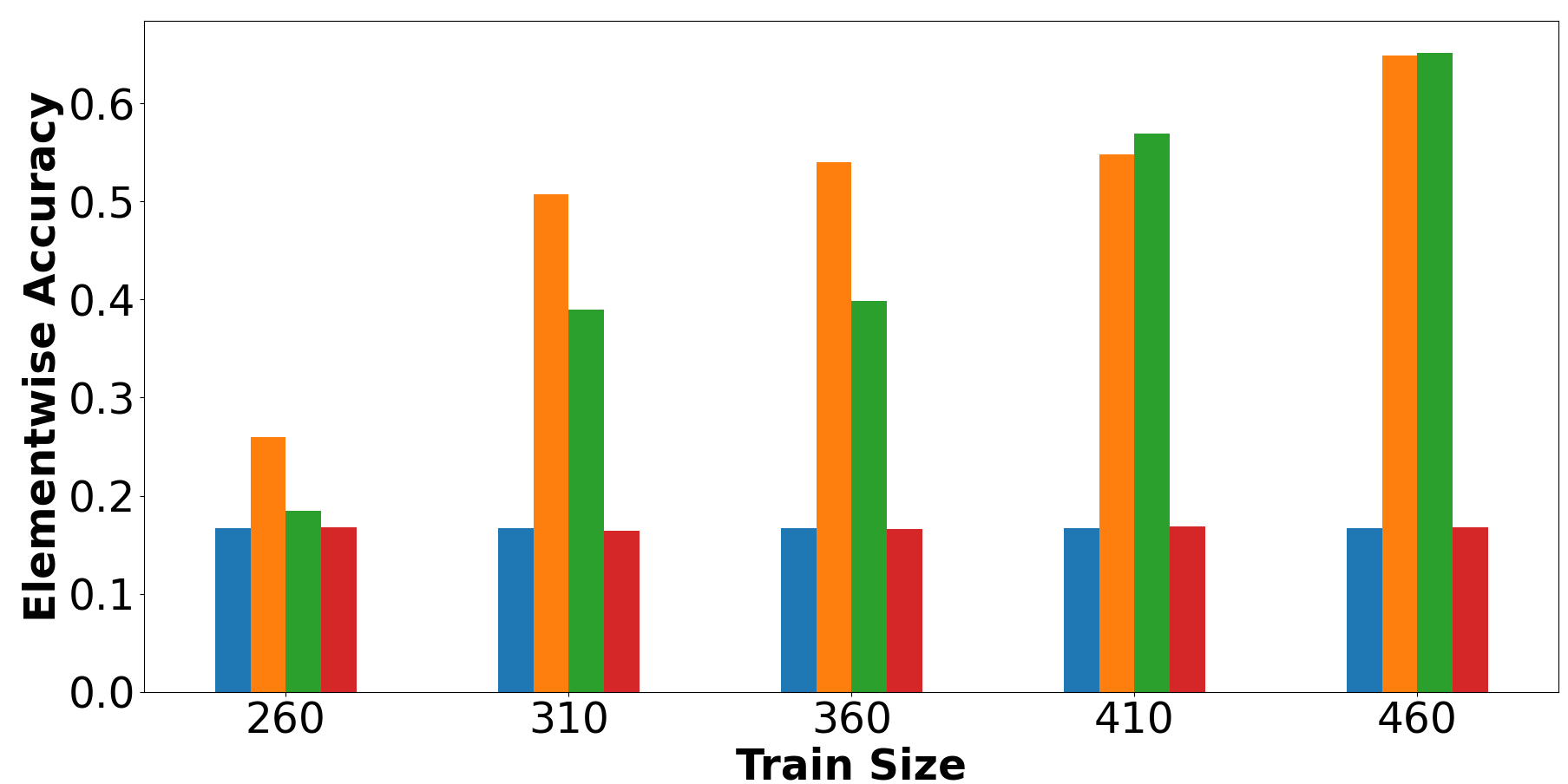}
	\caption{\centering Performance of RESOLVE compared to baselines for 6 elements sequence sorting. %CorelNet uses softmax activation.
    }
    \label{fig:sorting_result}
    \vspace{-0.5cm}
\end{wrapfigure}
We generate have generated random objects for the sorting task. First, we create two sets of random attributes: $\mathcal{A} = {a_1, a_2, a_3, a_4}$, where %$a_i \overset{iid}{\sim} \mathcal{N}(0, I) \in \mathbb{R}^{4}$, 
and $\mathcal{B} = {b_1, \ldots, b_{12}}$. %where $b_i \overset{iid}{\sim} \mathcal{N}(0, I) \in \mathbb{R}^{8}$. 
Each set of attributes has a strict ordering: $a_1 \prec a_2 \prec a_3 \prec a_4$ for $\mathcal{A}$ and $b_1 \prec b_2 \prec \cdots \prec b_{12}$ for $\mathcal{B}$. Our random objects are formed by taking the Cartesian product of these two sets, $\mathcal{O} = \mathcal{A} \times \mathcal{B}$, resulting in $N = 48$ objects. Each object in $\mathcal{O}$ is a vector in $\mathbb{R}^{12}$, formed by concatenating one attribute from $\mathcal{A}$ with one attribute from $\mathcal{B}$.

We then establish a strict ordering relation for $\mathcal{O}$, using the order relation of $\mathcal{A}$ as the primary key and the order relation of $\mathcal{B}$ as the secondary key. Specifically, $(a_i, b_j) \prec (a_k, b_l)$ if $a_i \prec a_k$ or if $a_i = a_k$ and $b_j \prec b_l$. We generated a randomly permuted set of 5 and a set of 6 objects in $\mathcal{O}$. The target sequences are the indices representing the sorted order of the object sequences (similar to the 'argsort' function). The training data is uniformly sampled from the set of 6 elements based sequences in $\mathcal{O}$. We generate non-overlapping validation and testing datasets in the following proportion: $20\%$ testing, $10\%$ validation and $70\%$ training.

We used \textit{element wise accuracy} to assess the performance of \texttt{RESOLVE}, as in \citep{abstractor}. The accuracy of \texttt{RESOLVE} is compared against the Relational Abstractor \citep{abstractor}, Transformer~\cite{vaswani2017attention} and CorelNet\citep{kerg2022neural}.

%\begin{figure}[ht]
%    \centering    \includegraphics[width=0.75\textwidth]{legend_sort.png}
%    \begin{subfigure}[t]{0.45\textwidth}
%        \centering\captionsetup{width=.9\linewidth}
%        % \vskip-20pt
%        \includegraphics[width=\textwidth]{5-sort.png}
%        \vskip-5pt
%        \caption{5 Elements Sequence Sorting}\label{fig:s-5}
%    \end{subfigure}
%    \begin{subfigure}[t]{0.45\textwidth}
%        \centering\captionsetup{width=.9\linewidth}
%        % \vskip-20pt
%        \includegraphics[width=\textwidth]{6-sort.png}
%        \vskip-5pt
%        \caption{6 Elements Sequence Sorting}\label{fig:s-6}% The Abstractor is nearly as sample 
%    \end{subfigure}
%    \caption{Performance of RESOLVE compared to baselines. CorelNet uses softmax activation.}
%    \vskip-10pt
%\end{figure}

%As the number of elements to sort increases, the performance of all models decreases slightly due to the increasing task complexity. 
Figure~\ref{fig:sorting_result} shows that \texttt{RESOLVE} achieves better accuracy than the baselines %($1.51$x to $2.24$x better than Relational-Abstractor for 5 elements and 
($1.56$x to $1.02$x better than Relational-Abstractor). Relational-Abstractor \citep{abstractor} still outperforms the transformer and CorelNet-Softmax, validating the results of \citep{abstractor}. \texttt{RESOLVE} demonstrates a high generalizability compared to SOTA. However, as the number of training sample increases, Relational Abstractor and \texttt{RESOLVE} converge toward the same level of accuracy with increased training data. %This shows the efficiency of the relational bottleneck used for explicit relational representation in both models. 

\subsection{Math problem-solving: partially-relational sequence-to-sequence tasks}

\begin{figure}[ht]
    \vskip-5pt
    \begin{center}
    \begin{small}
    \resizebox{\textwidth}{!}
    {
    \begin{tabular}{cc}
        \begin{tabular}{l}
        Task: \texttt{Numbers\_place\_value}\\
        Question: \texttt{what is the tens digit of 3585792?}\\
        Answer: \texttt{9}
        \end{tabular}
        &
        \begin{tabular}{l}
        Task: \texttt{Comparison\_pair}\\
        Question: \texttt{Which is bigger: 4/37 or 7/65?}\\
        Answer: \texttt{4/37 is bigger}
        \end{tabular}
    \end{tabular}
    }
    \end{small}
    \end{center}
    % \vskip-10pt
    \vspace{-0.1in}
    \caption{Examples of input/target sequences from the math problem-solving dataset.}\label{fig:math_dataset2}
    % \vskip-15pt
    \vspace{-0.5cm}
\end{figure}

% Please add the following required packages to your document preamble:
% \usepackage{multirow}
% \usepackage[table,xcdraw]{xcolor}
% Beamer presentation requires \usepackage{colortbl} instead of \usepackage[table,xcdraw]{xcolor}

\begin{table*}[htbp]
\begin{center}

\resizebox{\textwidth}{!}
{
\begin{tabular}{cc|cccc|cccc|cccc|c}
\cline{3-14}
 &  & \multicolumn{4}{c|}{Comparison Closest} & \multicolumn{4}{c|}{Comparison Pair} & \multicolumn{4}{c|}{Comparison Place Value} &  \\ \cline{2-15} 
\multicolumn{1}{c|}{}  & \textit{Train size} & \textit{100} & \multicolumn{1}{c}{\textit{1000}} & \textit{10000} & \textit{avg} & \textit{100} & \textit{1000} & \textit{10000} & \textit{avg} & \textit{100} & \textit{1000} & \textit{10000} & \textit{avg} & \multicolumn{1}{c|}{\textit{Overall}} \\ \hline
\multicolumn{1}{|c|}{} & RESOLVE & \textbf{15.08} & \textbf{20.88} & {\color[HTML]{000000} 52.36} & \textit{\textbf{29.44}} & {\color[HTML]{000000} \textit{28.43}} & \textbf{38.15} & 66.84 & \textit{\textbf{44.47}} & 19.36 & \textbf{36.98} & 98.68 & \textit{\textbf{51.67}} & \multicolumn{1}{c|}{\textbf{41.86}} \\ \cline{2-15} 
\multicolumn{1}{|c|}{} & Rel-Abstractor & 13.77 & 19.49 & \textbf{52.46} & \textit{28.57} & 26.86 & 35.82 & \textbf{69.19} & \textit{43.95} & 19.93 & 32.21 & 99.43 & \textit{50.52} & \multicolumn{1}{c|}{41.01} \\ \cline{2-15} 
\multicolumn{1}{|c|}{\multirow{-3}{*}{Model}} & Transformer & 14.25 & 18.08 & 37.76 & \textit{23.36} & 30.73 & 34.1 & 64.37 & \textit{43.06} & 21.1 & 32.91 & \textbf{99.64} & \textit{51.21} & \multicolumn{1}{c|}{39.21} \\ \hline
\end{tabular}}
\end{center}
\vspace{-0.1in}
\caption{\small \centering Accuracy (probability of correct answer) of \texttt{RESOLVE} compared to SOTA for three mathematical reasoning tasks (\textit{best accuracy in bold})}
\label{tab:math_res}
\vspace{-0.25cm}
\end{table*}

%The object-sorting experiments in Section \ref{object_sorting} are "purely relational" because the set of pairwise $\prec$ order relations alone provides sufficient information to solve the task. 
We further evaluate \texttt{RESOLVE} on a mathematical reasoning dataset (Figure~\ref{fig:math_dataset2}), which represents a \textit{partially relational} sequence-to-sequence problem. Table~\ref{tab:math_res} presents the accuracy achieved by the relational abstractor, transformer, and \texttt{RESOLVE} on three different datasets using 100 to 10,000 training samples. The accuracy corresponds to the percentage of full sequence matches, each one representing a correct answer. We report the average accuracy across the three different training sizes in the table, as well as an overall accuracy for all test cases. \texttt{RESOLVE} outperforms the state-of-the-art (SOTA) on average across the three test cases. It also achieves higher accuracy with a small training set, demonstrating the generalizability of the proposed architecture. Notably, neither the relational representation nor the object-level features alone are sufficient for inducing abstract rules from partially relational tasks, which penalizes both the Abstractor and transformer. In contrast, \texttt{RESOLVE} combines both levels of knowledge into a single structure.
\vspace{-0.05in}
\subsection{Computational Overhead Assessment}

\begin{table}[htpb]
\begin{center}
\resizebox{0.6\textwidth}{!}
{
\begin{tabular}{|c|ccc|ccc|}
\hline
Embedding size & \multicolumn{3}{c|}{\textit{32}} & \multicolumn{3}{c|}{\textit{64}} \\ \hline
Model & \multicolumn{1}{c|}{$\pi$} & \multicolumn{1}{c|}{$\beta$ (L1 Cache)} & $\beta$ (DRAM) & \multicolumn{1}{c|}{$\pi$} & \multicolumn{1}{c|}{$\beta$ (L1 Cache)} & $\beta$ (DRAM) \\ \hline
HD Attention & \multicolumn{1}{c|}{\textbf{1.99}} & \multicolumn{1}{c|}{\textbf{0.787}} & 0.867 & \multicolumn{1}{c|}{\textbf{1.99}} & \multicolumn{1}{c|}{\textbf{0.788}} & \textbf{0.870} \\ \hline
Self-Attention & \multicolumn{1}{c|}{1.99} & \multicolumn{1}{c|}{0.783} & \textbf{0.869} & \multicolumn{1}{c|}{1.97} & \multicolumn{1}{c|}{0.781} & \textbf{0.863} \\ \hline
\end{tabular}
}
\end{center}
\vspace{-0.1in}
\caption{\small \centering Comparison between the HD-Attention Self attention mechanism in term of computational overhead. $\beta$ is the bandwidth bound in \textit{Flop/Byte} and $\pi$ is processor peak performance in \textit{GFLOPS}}
\vspace{-0.1in}
\label{tab:overhead_roofline}
\end{table}

We assessed the computational overhead of the HD-Attention mechanism described in Section~\ref{resolve_bundling} against the baseline of a regular self attention mechanism~\cite{vaswani2017attention}. The operations are done on a CPU using Multi-threading. The memory overhead is measured at the level of DRAM and L1 cache memory using the roofline model~\cite{ofenbeck2014applying}. A high $\beta$ value means that the system is less likely to encounter memory bottlenecks. A high $\pi$ means the processor is capable of performing more computations per cycle. Table~\ref{tab:overhead_roofline} shows that the HD-Attention mechansim has better computational performance compared to self-attention ($\pi$) with higher memory bandwidth $\beta$. This  is due to the use of the bipolar HD representation and operations such as bundling and binding.     

\section{Conclusion}
In this work we have presented \texttt{RESOLVE}, a vector-symbolic framework for relational learning that outperforms the state of the art thanks to its use of high-dimensional attention mappings for mixing relational representations and object features. In future we plan to examine multimodal learning tasks and sequence-to-sequence learning tasks in the high-dimensional domain, taking advantage of the computational efficiency of vector-symbolic architectures.

\bibliography{iclr2025_conference}
\bibliographystyle{iclr2025_conference}

\appendix
\section{Appendix}
\subsection*{Code and Reproducibility} The code, detailed experimental logs, and instructions for reproducing our experimental results are available at: \url{https://github.com/mmejri3/RESOLVE}.

\subsection*{Single Output Tasks} In this section, we provide comprehensive information on the architectures, hyperparameters, and implementation details of our experiments. All models and experiments were developed using TensorFlow. The code, along with detailed experimental logs and instructions for reproduction, is available in the project's public repository.

\subsection{Computational Resources} For training the RESOLVE and SOTA models on the single-output relational tasks, we used a GPU (Nvidia RTX A6000 with 48GB of RAM). For training LARS-VSA and SOTA on purely and partially sequence-to-sequence abstract reasoning tasks, we used a single GPU (Nvidia A100 with 80GB of RAM). The overhead assessment of the HD-attention mechanism was conducted on a CPU (11th Gen Intel® Core™ i7).

\subsection{\textit{Single Output Purely Relational Tasks}}

\subsubsection{Pairwise Order} Each model in this experiment follows the structure: \texttt{input} $\to$ {\texttt{module}} $\to$ \texttt{flatten} $\to$ \texttt{MLP}, where {\texttt{module}} represents one of the described modules, and \texttt{MLP} is a multilayer perceptron with one hidden layer of 32 neurons activated by ReLU.

\paragraph{RESOLVE Architecture} Each model consists of a single module and a hypervector of dimensionality $D = 1024$. We use a dropout rate of 0.1 to prevent overfitting. The two hypervectors are flattened and passed through hidden layers containing 32 neurons with ReLU activation, followed by a final layer with one neuron activated by a sigmoid function.

\paragraph{Abstractor Architecture} The Abstractor module utilizes the following hyperparameters: number of layers $L = 1$, relation dimension $d_r = 4$, symbol dimension $d_s = 64$, projection (key) dimension $d_k = 16$, feedforward hidden dimension $d_{\mathrm{ff}} = 64$, and relation activation function $\sigma_{\mathrm{rel}} = \mathrm{softmax}$. No layer normalization or residual connections are applied. Positional symbols, which are learned parameters, are used as the symbol assignment mechanism. The output of the Abstractor module is flattened and passed to the \texttt{MLP}.

\paragraph{CoRelNet Architecture} CoRelNet has no hyperparameters. Given a sequence of objects, $X = (x_1, \ldots, x_m)$, standard CoRelNet~\citep{kerg2022neural} computes the inner product and applies the Softmax function. We also add a learnable linear map, $W \in \mathbb{R}^{d \times d}$. Hence, $\bar{R} = \text{Softmax}(R)$, where $R = \left[\langle W x_i, W x_j\rangle\right]{ij}$. The CoRelNet architecture flattens $\bar{R}$ and passes it to an \texttt{MLP} to produce the output. The asymmetric variant of CoRelNet is given by $\bar{R} = \text{Softmax}(R)$, where $R = \left[\langle W_1 x_i, W_2 x_j\rangle\right]{ij}$, and $W_1, W_2 \in \mathbb{R}^{d \times d}$ are learnable matrices.

\paragraph{PrediNet Architecture} Our implementation of PrediNet~\citep{shanahan2020explicitly} is based on the authors' publicly available code. We used the following hyperparameters: 4 heads, 16 relations, and a key dimension of 4 (see the original paper for the definitions of these hyperparameters). The output of the PrediNet module is flattened and passed to the \texttt{MLP}.

\paragraph{MLP} The embeddings of the objects are concatenated and passed directly to an \texttt{MLP}, which has two hidden layers, each containing 32 neurons with ReLU activation.

\paragraph{Training/Evaluation} We use cross-entropy loss and the AdamW optimizer with a learning rate of $10^{-4}$. The batch size is 128, and training is conducted for 100 epochs. Evaluation is performed on the test set. The experiments are repeated 5 times, and we report the mean accuracy and standard deviation.

\subsubsection{\textit{SET}}

The card images are RGB images with dimensions of $70 \times 50 \times 3$. A CNN embedder processes these images individually, producing embeddings of dimension $d=64$ for each card. The CNN is trained to predict four attributes of each card. After training, embeddings are extracted from an intermediate layer, and the CNN parameters are frozen. The common architecture follows the structure: \texttt{CNN Embedder} $\to$ {\texttt{Abstractor, CoRelNet, PrediNet, MLP}} $\to$ \texttt{Flatten} $\to$ \texttt{Dense(2)}. Initial tests with the standard CoRelNet showed no learning. However, removing the Softmax activation improved performance slightly. Hyperparameter details are provided below.

\textbf{Common Embedder Architecture:} The architecture follows this structure: \texttt{Conv2D} $\to$ \texttt{MaxPool2D} $\to$ \texttt{Conv2D} $\to$ \texttt{MaxPool2D} $\to$ \texttt{Flatten} $\to$ \texttt{Dense(64, ReLU)} $\to$ \texttt{Dense(64, ReLU)} $\to$ \texttt{Dense(2)}. The embedding is taken from the penultimate layer. The CNN is trained to perfectly predict the four attributes of each card, achieving near-zero loss.

\textbf{RESOLVE Architecture:} The RESOLVE module has the following hyperparameters: hypervector dimension $D = 1024$. The outputs are flattened and passed through a feedforward hidden layer with dimension $d_{\mathrm{ff}} = 64$, followed by a final layer with a single neuron and sigmoid activation. A dropout rate of 0.4 is used to prevent overfitting.

\textbf{Abstractor Architecture:} The Abstractor module uses the following hyperparameters: number of layers $L = 1$, relation dimension $d_r = 4$, symmetric relations ($W_q^{i} = W_k^{i}$ for $i \in [d_r]$), ReLU activation for relations, symbol dimension $d_s = 64$, projection (key) dimension $d_k = 16$, feedforward hidden dimension $d_{\mathrm{ff}} = 64$, and no layer normalization or residual connections. Positional symbols, which are learned parameters, are used as the symbol assignment mechanism.

\textbf{CoRelNet Architecture:} In this variant of CoRelNet, we found that removing the Softmax activation improved performance. The standard CoRelNet computes $R = \text{Softmax}(A)$, where $A = \left[\langle W x_i, W x_j\rangle\right]_{ij}$. 

\textbf{PrediNet Architecture:} The hyperparameters used are 4 heads, 16 relations, and a key dimension of 4, as described in the original paper. The output of the PrediNet module is flattened and passed to the MLP.

\textbf{MLP:} The embeddings of the objects are concatenated and passed directly to an MLP with two hidden layers, each containing 32 neurons with ReLU activation.

\textbf{Data Generation:} The dataset is generated by randomly sampling a "set" with probability 1/2 and a non-"set" with probability 1/2. The triplet of cards is then randomly shuffled.

\textbf{Training/Evaluation:} We use cross-entropy loss and the AdamW optimizer with a learning rate of $10^{-4}$. The batch size is 512, and training is conducted for 200 epochs. Evaluation is performed on the test set. We train our model on a randomly sampled set of $N$ samples, where $N \in {500, 700, 900, 1100, 1300, 1500, 1700}$.
\subsection{\textit{Single Output Partially Relational Tasks}}

\subsubsection{\textit{SET}}

We used the same settings as in the previous SET experiment. However, in this task, the input features used as a sequence of objects are derived from the first convolutional layer of the pre-trained \texttt{CNN}. This approach avoids using highly processed object-level features, allowing us to assess the ability of RESOLVE and the baseline models to capture both object-level features and relational representations.

We did not change the hyperparameters of the baseline models or the RESOLVE model. However, we added an attentional encoder at the front end, with a single layer and two heads.

\subsubsection{\textit{MNIST-MATH}}

This experiment is inspired by the MNIST digits addition task introduced by~\cite{manhaeve2018deepproblog}. We randomly selected 10,000 pairs of MNIST digits from the MNIST training set and generated labels using a non-linear mathematical formula: $F(a, b) = \lvert 3a - 2b \rvert$.

The digits are normalized and flattened before being passed to the relational models. We used the same hyperparameters as in the SET experiment.

\subsection{Relational Sequence-to-Sequence Tasks}

\subsubsection{\textit{Object-Sorting Task}}

\paragraph{RESOLVE Architecture} We used architecture (c) from Figure \ref{fig:architecture}. The encoder includes a BatchNormalization layer. The RESOLVE architecture consists of a single module with a hyperdimensional dimension of $D = 1024$. The decoder has 4 layers, 2 attention heads, a feedforward network with 64 hidden units, and a model dimension of 64.

\paragraph{Abstractor Architecture} Each of the Encoder, Abstractor, and Decoder modules consists of $L = 2$ layers, with 2 attention heads/relation dimensions, a feedforward network with $d_{\mathrm{ff}} = 64$ hidden units, and a model/symbol dimension of $d_{\mathrm{model}} = 64$. The relation activation function is $\sigma_{\mathrm{rel}} = \mathrm{Softmax}$. Positional symbols are used as the symbol assignment mechanism, which are learned parameters of the model.

\paragraph{Transformer Architecture} We implemented the standard Transformer architecture as described by~\citep{vaswani2017attention}. Both the Encoder and Decoder modules share the same hyperparameters, with an increased number of layers. Specifically, we use 4 layers, 2 attention heads, a feedforward network with 64 hidden units, and a model dimension of 64.

\paragraph{Training and Evaluation} The models are trained using cross-entropy loss and the Adam optimizer with a learning rate of $5 \cdot 10^{-4}$. We use a batch size of 128 and train for 500 epochs. To evaluate the learning curves, we vary the training set size, sampling random subsets ranging from 260 to 460 samples in increments of 50. Each sample consists of an input-output sequence pair. For each model and training set size, we perform 10 runs with different random seeds and report the mean accuracy.

\subsubsection{\textit{Math Problem-Solving}}

The dataset consists of various math problem-solving tasks, each featuring a collection of question-answer pairs. These tasks cover areas such as solving equations, expanding polynomial products, differentiating functions, predicting sequence terms, and more. The dataset includes 2 million training examples and 10,000 validation examples per task. Questions are limited to a maximum length of 160 characters, while answers are restricted to 30 characters. Character-level encoding is used, with a shared alphabet of 95 characters, which includes upper and lower case letters, digits, punctuation, and special tokens for start, end, and padding.

\paragraph{Abstractor Architectures} The Encoder, Abstractor, and Decoder modules share identical hyperparameters: number of layers $L = 1$, relation dimension/number of heads $d_r = n_h = 2$, symbol dimension/model dimension $d_s = d_{\mathrm{model}} = 64$, projection (key) dimension $d_k = 32$, and feedforward hidden dimension $d_{\mathrm{ff}} = 128$. The relation activation function in the Abstractor is $\sigma_{\mathrm{rel}} = \mathrm{Softmax}$. One model uses positional symbols with sinusoidal embeddings, while the other uses symbolic attention with a symbol library of $n_s = 128$ learned symbols and 2-head symbolic attention.

\paragraph{Transformer Architecture} The Transformer Encoder and Decoder have the same hyperparameters as the Encoder and Decoder in the Abstractor architecture.

\paragraph{RESOLVE Architectures} The RESOLVE model follows architecture (D) from Figure \ref{fig:architecture}. We use the same Decoder as the Abstractor architecture. The RESOLVE model has a single module and a hyperdimensional dimension of $D = 1024$.

\paragraph{Training and Evaluation} Each model is trained for 1000 epochs using categorical cross-entropy loss and the Adam optimizer with a learning rate of $6 \times 10^{-4}$, $\beta_1 = 0.9$, $\beta_2 = 0.995$, and $\varepsilon = 10^{-9}$. The batch size is 64. The training set consists of $N$ samples, where $N \in {100, 1000, 10,000}$.

\end{document}